\pdfoutput=1
\documentclass[11pt]{article}

\usepackage{url}
\usepackage{subfigure}
\usepackage{multirow}
\usepackage{enumitem}
\usepackage{stmaryrd}
\usepackage{array}
\usepackage{threeparttable}
\usepackage{blindtext}
\usepackage{amssymb}

\usepackage[ruled,vlined,linesnumbered,longend]{algorithm2e}
\usepackage{svg}
\usepackage{placeins}

\usepackage{soul}
\usepackage[utf8]{inputenc}
\usepackage{graphicx}
\usepackage{amsmath}
\usepackage{booktabs}
\usepackage{lipsum}

\usepackage{threeparttable}
\usepackage{makecell}
\usepackage{booktabs}
\usepackage{multirow}
\usepackage{marvosym}
\usepackage{tikz}
\newcommand{\hide}[1]{}
\newcommand{\vpara}[1]{\vspace{0.02in}\noindent\textbf{#1}}

\newcommand{\model}{DPSG}

\newcommand*\emptycirc[1][1ex]{\tikz\draw (0,0) circle (#1);} 
\newcommand*\halfcirc[1][1ex]{%
	\begin{tikzpicture}
	\draw[fill] (0,0)-- (270:#1) arc (270:450:#1) -- cycle ;
	\draw (0,0) circle (#1);
	\end{tikzpicture}}
\newcommand*\fullcirc[1][1ex]{\tikz\fill (0,0) circle (#1);} 

\newcommand\ring[1][1ex]{
    \tikz \draw (0,0) circle (#1)
                (0,0) circle (#1-1);
}
\usepackage{amsmath,amsfonts,bm}

\def\eqref#1{equation~\ref{#1}}
\def\1{\bm{1}}

\def\rvo{{\mathbf{o}}}

\def\rvs{{\mathbf{s}}}
\def\rvt{{\mathbf{t}}}

\def\rvx{{\mathbf{x}}}
\def\rvy{{\mathbf{y}}}

\def\rmL{{\mathbf{L}}}
\def\rmM{{\mathbf{M}}}

\def\rmP{{\mathbf{P}}}

\def\rmS{{\mathbf{S}}}

\DeclareMathAlphabet{\mathsfit}{\encodingdefault}{\sfdefault}{m}{sl}
\SetMathAlphabet{\mathsfit}{bold}{\encodingdefault}{\sfdefault}{bx}{n}

\def\sR{{\mathbb{R}}}

\def\sV{{\mathbb{V}}}

\usepackage{acl}

\usepackage{times}
\usepackage{latexsym}

\usepackage[T1]{fontenc}
\usepackage[utf8]{inputenc}

\usepackage{microtype}

\title{
Schema-Free Dependency Parsing via Sequence Generation
}

\author{
    Boda Lin$^{1\ddagger}$, Zijun Yao$^{2\ddagger}$, Jiaxin Shi$^3$, Shulin Cao$^2$, Binghao Tang$^1$, Si Li$^{1*}$, Yong Luo$^4$, Juanzi Li$^2$ and Lei Hou$^2$ \\
    $^1$School of Artificial Intelligence, Beijing University of Posts and Telecommunications\\
    $^2$Department of Computer Science and Technology, Tsinghua University\\
    $^3$Huawei Cloud Computing Technologies
    $^4$School of Computer Science, Wuhan University \\
  \texttt{\{linboda, lisi\}@bupt.edu.cn} \\
  \texttt{\{yaozj20@mails., houlei@\}tsinghua.edu.cn}
   }

\begin{document}
\maketitle

\begin{abstract}
Dependency parsing aims to extract syntactic dependency structure or semantic dependency structure for sentences.
Existing methods suffer the drawbacks of lacking universality or highly relying on the auxiliary decoder.
To remedy these drawbacks, we propose to achieve universal and schema-free Dependency Parsing (DP) via Sequence Generation (SG) DPSG by utilizing only the pre-trained language model (PLM) without any auxiliary structures or parsing algorithms.
We first explore different serialization designing strategies for converting parsing structures into sequences. Then we design \textit{dependency units} and concatenate these units into the sequence for DPSG.
Thanks to the high flexibility of the sequence generation, our DPSG can achieve both syntactic DP and semantic DP using a single model.
By concatenating the prefix to indicate the specific schema with the sequence, our DPSG can even accomplish multi-schemata parsing.
The effectiveness of our DPSG is demonstrated by the experiments on widely used DP benchmarks, i.e., PTB, CODT, SDP15, and SemEval16.
DPSG achieves comparable results with the first-tier methods on all the benchmarks and even the state-of-the-art (SOTA) performance in CODT and SemEval16.
This paper demonstrates our DPSG has the potential to be a new parsing paradigm.
We will release our codes upon acceptance.
\let\thefootnote\relax\footnotetext{$^\ddagger$Boda Lin and Zijun Yao make equal contribution}
\let\thefootnote\relax\footnotetext{$^*$Corresponding author}
\end{abstract}

\section{Introduction}

\begin{figure}[t]
\centering
    \includegraphics[width=1.0\linewidth]{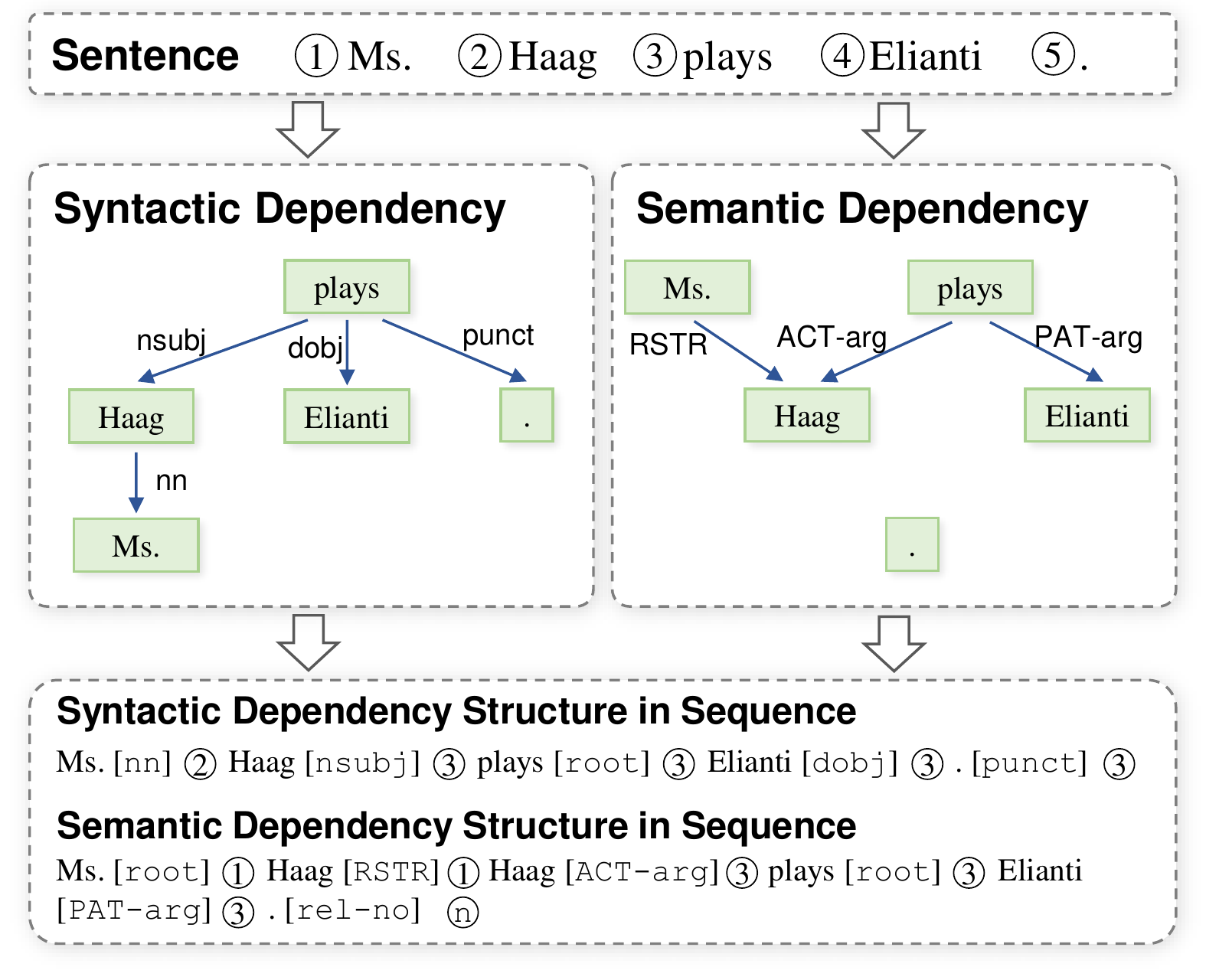}
% \vspace{-0.3in}
\caption{
% This sentence "Ms. Haag plays Elianti ." is from Penn Treebank3. StanFord means the syntactic parsing tree convert from StanfordNLP. DM, PAS and PSD are three different schema of semantic dependency graph.
% Two different dependency structures of the same sentence ``\textit{Ms. Haag plays Elianti .}''.
% The presented syntactic dependency structure adopts the Stanford schema~\cite{manning2014stanford} and the semantic dependency structure follows the PSD schema~\cite{stephan2015sdp15}.
Parsing ``\textit{Ms. Haag plays Elianti .}'' according to the Stanford syntactic dependency structure~\cite{manning2014stanford} and the PSD semantic dependency structure~\cite{stephan2015sdp15}.
They are further converted into unified serialized representations.
}
\label{fig:parsingtree}
\end{figure}

% \emph{syntactic} or \emph{semantic}
Dependency Parsing (DP), which aims to extract the structural information beneath sentences, is fundamental in understanding natural languages.
It benefits a wide range of Natural Language Processing~(NLP) applications, such as 
% information extraction~\cite{attardi2014ie}, 
machine translation~\cite{bugliarello2020mt}, 
question answering~\cite{teney2017qa}, 
and information retrieval~\cite{chandurkar2017ir}.
As shown in Figure~\ref{fig:parsingtree}, dependency parsing predicts for each word the existence and dependency relation with other words according to a pre-defined schema.
Such dependency structure is represented in tree or directed acyclic graph, which can be converted into flattened sequence, as presented in this paper.

% 应该再增加一句，与图片对应，说明如图所示，我们可以直观地看到，这两类结构是可以被序列化的，这样在开头就点出了工作的重点

% Recently, there are two main threads in DP studies.
% One thread further subdivides DP into more specific tasks or optimizes DP in various data-scarce situation. 
% \citet{someone} start to consider semantic dependency relation beyond the syntactic dependency relation.
% \citet{someone} highlight the importance of language-specific features.
% % and introduce cross-lingual DP task.
% \citet{someone} argue that dependency relation should be domain-invariant and propose cross-domain DP task.
% The other thread engineers more sophisticated parsing modules. 
% With the promising external knowledge encoded by Pre-trained Language Models (PLM), a large amount of studies~\cite{something} successfully verifies the effectiveness of incorporating contextual embeddings~\cite{bert, roberta}.
% There are also efforts to leverage cross-domain (-lingual) dependency knowledge with transfer learning techniques.
% These two threads are inter-winded and motivate the development of each other, 

% % lin
% However, most previous research for DP focus on transition-based methods and graph-based methods.
% Altougth outstanding performance, these methods also need the auxiliary traditional parsing algorithms, which make the parsing process cubersome.
% %% 变得越来越繁琐
% On the other hand, the usage of PLM is narrowd in replacing the original embedding layer to get a better word representation.

% In this paper, we propose treat dependency parsing as sequence generation.
% By Serializing parsing structure, the workflow of DP change into a seq-2-seq paradigm.
% The input of our method is the original sequence, and the output is the serialization of the parsing structure.
% To get the high-quality serialization sequence and fully utilize the potential of PLM, we use T5 and mT5 as the backbone to generate the parsing sequence.

% \boda{do not divide the subtasks, new directions introduce by 1, 2, 3}
% Previous dependency parsing mainly focus on develop the performance of three mainstreams methods
The field of dependency parsing develops three main categories of paradigms: graph-based methods~\cite{dozat2017biaffine}, transition-based methods~\cite{ma2018stackptr}, and sequence-based methods~\cite{li2018seq2seq}.
While prospering with these methods, dependency parsing shows three trends now. 
1) New Schema. Recent works extend dependency parsing from syntactic DP (SyDP) to semantic DP (SeDP) with many new schemata~\cite{stephan2015sdp15,che2016semeval}. 
2) Cross-Domain. Corpora from different domains facilitate the research on cross-domain dependency parsing~\cite{peng2019nlpcc,li2019codt}.
3) PLM. With the development of pre-trained language models~(PLMs), researchers manage to enable PLMs on dependency task and successfully achieve the new state-of-the-art (SOTA) results~\cite{fernandez2020transition,gan21mrc}.
However, there are still two main issues.

\textbf{Lacking Universality.}
Although there are many successful parsers, most of them are schema-specific and have limitations,
e.g., sequence-based parsers~\cite{vacareanu2020pat} are only suitable for SyDP.
% For example, machine reading comprehension based parsers~\cite{gan21mrc} are only suitable for syntactic dependency parsing.
Thus, these methods require re-training before being adapted to another schema.
% which is extremely costly for the PLM based parsers.
% Both parsers with PLMs and without PLMs can achieve outstanding performances on parsing task.
% But most of them are schema-speific models.
% \boda{Need more details, such as parameters}.
% In addition, most previous parsers require auxiliary pseudo-labeling processes or domain adaptation modules for unsupervised cross-domain parsing~\cite{yu2015self,peng2019nlpcc,lin2021unsupervised}.
% \boda{I will add citations later}

\textbf{Relying on Extra Decoder.}
Previous parsers usually produce the parsing results employing an extra decoding module, such as a biaffine network for score calculation~\cite{dozat2017biaffine} and a neural transducer for decision making~\cite{zhang2019broad}.
These modules cannot be pre-trained and learn the dependency relation merely from the training corpora. 
Thus, only part of these models generalizes to sentences of different domains.

To address these issues, 
we propose schema-free \textbf{D}ependency \textbf{P}arsing via \textbf{S}equence \textbf{G}eneration (\model).
% and use a single encoder-decoder pre-trained language model to directly generate the dependency structure as a sequence.
The core idea is to find a unified unambiguous serialized representation for both syntactic and semantic dependency structures.
Then an encoder-decoder PLM is learned to generate the parsing results following the serialized representation, without the need for an additional decoder.
That is, our parser can achieve its function using one original PLM (without any modification), and thus is entirely pre-trained.
% then learns to generate the parsing results as the serialized representation, without the need for additional decoder.
% In this case, the parser implements its functionality as a native PLM, and are thus completely pre-trained.
% Furthermore, the unified representation provides a principled way to pack different schemata into a single model.
Furthermore, by adding a prefix to the serialized representation, \model~provides a principled way to pack different schemata into a single model.
% The PLM takes on the role of dependency parser, and are completely pre-trained.
% The serialized representation directly encodes how many head words there are for each dependent word, rather than using a heuristic threshold to decide for each word whether it is a head word.
% In this way, a single encoder-decoder pre-trained language model can generate the parsing results in sequences alone, without the need for additional decoders.
% This prones potential bias towards the dependency knowledge in the training dataset domain.
% and could be less biased towards the dependency knowledge in the training dataset domain.

In particular, \model~consists of three key components.
The \textit{Serializer} is responsible for converting between the dependency structure and the serialized representation.
The \textit{Positional Prompt} pattern provides supplementary word position information in the input sentence to facilitate the sequence generation process. 
% ease the sequence generation process.
The \textit{encoder-decoder PLM} with added special tokens performs the parsing task via sequence generation.
The main advantages of \model~comparing with previous paradigms are summarized in Table~\ref{tab:intro:compare}.
Our \model~accomplishes DP for different schemata, unifies multiple schemata without training multiple models, and transfers the overall model to different domains.

We conduct experiments on $4$ popular DP benchmarks: PTB, CODT, SDP15, and SemEval16.
\model~performs generally well on different DP.
It significantly outperforms the baselines on cross-domain (CODT) and Chinese SeDP (SemEval16) corpora, and achieves comparable results on the other two benchmarks,
which further shows that our \model~has the potential to be a new paradigm for dependency parsing.
\begin{table}[t]
    \centering
    \scalebox{0.88}{
    \setlength{\tabcolsep}{3pt}
    \begin{tabular}{l|cccc}
    \toprule
        % Method & In-domain SyDP & In-domain SeDP & Cross-domain SyDP & Multi-Schema \\
        Paradigms & SyDP & SeDP & \begin{tabular}{c} Multi- \\ Schema \end{tabular} & \begin{tabular}{c} Unsupervised \\ Cross-Domain \end{tabular} \\
    \midrule
        Transition & {\small\fullcirc} & {\small\fullcirc} & {\small\emptycirc} & {\small\halfcirc} \\
        Graph & {\small\fullcirc} & {\small\ring} & {\small\emptycirc} & {\small\halfcirc} \\
        Sequence & {\small\fullcirc} & {\small\emptycirc} & {\small\emptycirc} & {\small\halfcirc} \\
        \model & {\small\fullcirc} & {\small\fullcirc} & {\small\fullcirc} & {\small\fullcirc} \\
    \bottomrule
    \end{tabular}
    }
    \caption{Summary of the previous parsing paradigms and \model.
    % ``SyDP", ``SeDP", ``Unsupervised Cross-Domain" and ``Multi-Schema" refer to different parsing scenarios. 
    {\small\fullcirc} means ``can be directly used in this scenario'', 
    {\small\ring}means ``can be used in this scenario after modification'',
    {\small\halfcirc}~means ``can partially generalize to this scenario'',
    and {\small\emptycirc} means ``cannot be used in this scenario''.}
    \label{tab:intro:compare}
\end{table}

\section{Preliminaries}

We formally introduce the dependency parsing task and the encoder-decoder PLM, and the corresponding notations.
This paper uses bold lower case letters, blackboard letters, and bold upper case letters to denote sequences, sets, and functions, respectively.
Elements in the sequence and the sets are enclosed in parentheses and braces, respectively.

\subsection{Dependency Parsing}

A pre-defined dependency schema is a set of relations $\sR$.
Dependency parsing takes a sentence 
% \begin{equation*}
$
\rvx = \left(w_1, w_2,..., w_n\right)
$
% \end{equation*}
as input, where $w_i$ is the $i^{\text{th}}$ word in the sentence.
It outputs the set of dependency pairs 
$\rvy = \left(p_1, p_2,..., p_n\right)$,
% \begin{equation*}
% \rvy = \left\{\left.\left(w_i, \left\{\left(h_i^j,r_i^j\right)\right\}\right) \right| w_i \in \rvx \right\}
% \end{equation*}
where $p_i = \left\{\left(r_i^j, h_i^j\right)\right\}$ denotes the dependency pair of the $i^\text{th}$ word $w_i$.
We use $h_i^j$ and $r_i^j$ to denote the $j^{\text{th}}$ head word of $w_i$ and their relation.
% $r_i^j \in \sR$ denotes the relation between $w_i$ and $h_i^j$.
$\mathtt{POS}(w)$ denotes the position of the specific word $w$ in the input sentence.
% The position of the head word in the input sentence is denoted as $\mathtt{POS}(h_i^j)$.
% Supposing $h_i^j$ refers to $w_k$, we have $\mathtt{POS}(h_i^j) = k$.
% Supposing $h_i^j = w_k$, we have $\mathtt{POS}(h_i^j) = k$.
% is the corresponding relation according to the schema.

% We first give the formal definition of dependency parsing:

% \vpara{Input:} $x = \{w_1, w_2,..., w_n\}$ is the input sentence, where $w_i$ is the $i^{\text{th}}$ word. 

% \vpara{Output:} y = $\{(w_i \xleftarrow{r_i}h(w_i))\}$ is the set of dependency pairs, where $h(w_i)$ is the head word of $w_i$ and $r_i$ is the relationship between $w_i$ and $h(w_i)$.

% Given a sentence ${w_0, w_1,..., w_n}$, the task of DP is predicting a set of dependencies pairs $\{(w_i \xleftarrow{r_i}h_i)\}$, which $w_i$ is the $i$-th token in sentence, $h_i$ is the head word of $i$-th token and $r_i$ is the dependency relation between the $w_i$ and $h_i$.

\begin{figure*}[t]
\centering
    \includegraphics[width=1.0\linewidth]{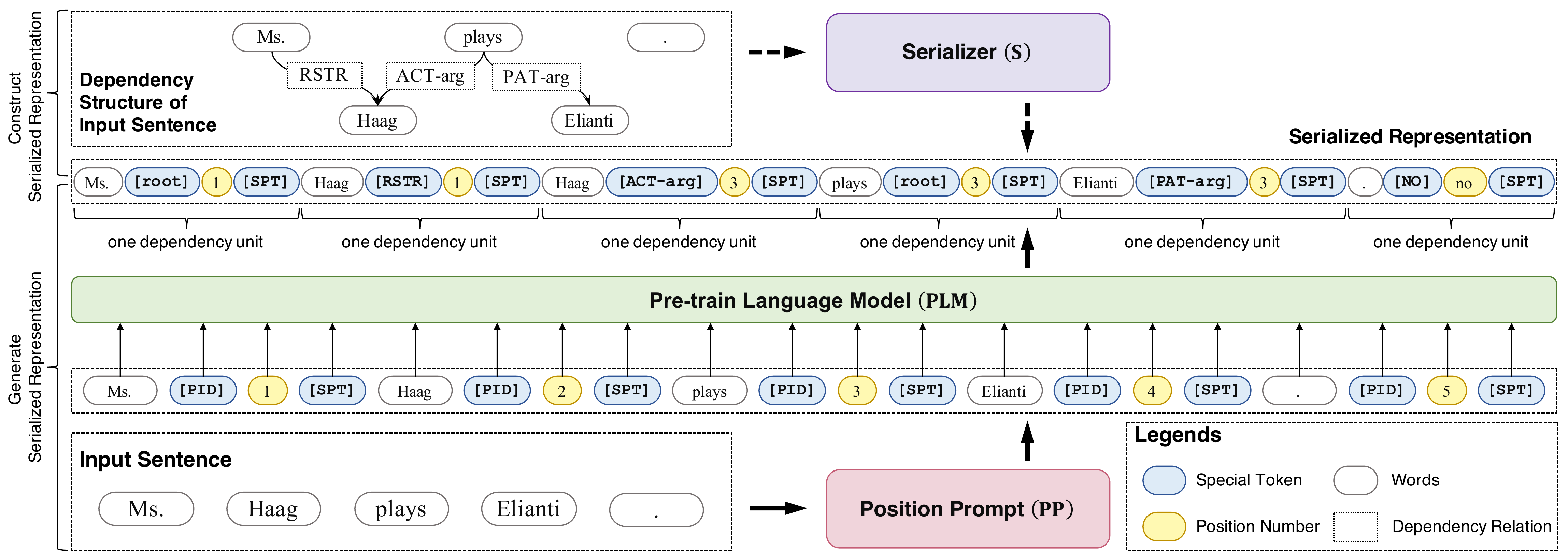}
\caption{
This figure shows the overall framework of \model.
The PSD semantic dependency structure of ``\textit{Ms. Haag plays Elianti .}'' is converted into the serialized representation by the Serializer.
The Positional Prompt module injects positional information into the input sentence, and the PLM is responsible for generating the results.
% ``\textit{Ms. Haag plays Elianti}'' is first fed into the Positional Prompt module.
% The input sentence with the positional information is sent into the PLM to generate the serialized representation, which is converted into the PSD semantic dependency structure with the (Inverse) Serializer.
}
\label{fig:model}
\end{figure*}

\textbf{Syntactic Dependency Parsing (SyDP)} 
analyses the grammatical dependency relations.
% Syntactic Dependency Parsing (SyDP) focuses more on the syntactic and shallow-semantic information of sentences.
% The relationships in SyDP often do not directly represent the semantic relationship between words, but are used to describe the grammatical and functional relationship, such as subject-predicate relation, verb-object relation, etc.
The parsing result of SyDP is a tree structure called the syntactic parsing tree. 
In the SyDP, each non-root word has exactly one head word, which means $\left| p_i \right| = 1$ if $w_i$ is the not root word.
% The formalization of SyDP is: given a sentence $\{w_0, w_1,..., w_n\}$, to predict the syntactic syntax tree $\{(w_i \xleftarrow{r_i}h_i)|i=0, 1,...n\}$.

\textbf{Semantic Dependency Parsing (SeDP)}
focuses on representing the deep-semantic relation between words. 
Each word in SeDP is allowed to have \textit{multiple} (even no) head words. 
This leads to the result of SeDP being a directed acyclic graph called Semantic Dependency Graph.
% The difference between SyDP and SeDP is illustrated in Figure~\ref{fig:parsingtree}.
Figure~\ref{fig:parsingtree} shows the difference between SyDP and SeDP, where SyDP produces a tree while SeDP produces a graph.
% which can be formalized as $\{(w_i \xleftarrow{}\{(h^{j}_i, r^{j}_i)|j \in \{0, 1,..., n-1\}\})|i=0, 1,..., n\}$.

% The formalization of Semantic Dependency Graph is $\{(w_i \xleftarrow{r_i}h_i)|i=0, 1,...n\}$.

% \vpara{Cross-domain Dependency Parsing.}
% Although DP can provide many useful information, the high cost of data labeling restricts the application of DP in downstream tasks. The labeling of DP requires the rich knowledge of linguistics, and the mainstream treebanks of DP in about the news and magazine. Due to the different semantic distribution and grammatical structure, the performance of DP dropped significantly when applied in other domains such as novel and web text. The Cross-domain Dependency Parsing (CDP), especially the unsupervised CDP is one of the most important sub-task of DP. In many scenarios, the unsupervised text is extremely easy to obtain, which means the key-point of CDP is how to utilize the unsupervised text of target domain.

\subsection{Pre-trained Language Model}
PLMs are usually stacks of attention blocks of Transformer~\cite{vaswani2017attention}.
% In this paper, we focus on PLMs with both encoder blocks and decoder blocks, such as T5~\cite{colin2020t5} and BART~\cite{lewis2020bart}.
% PLMs that are constructed by encoder blocks only are not feasible for sequence generation, and are thus not considered.
Some PLMs that consist of encoder blocks only (e.g., BERT~\cite{devlin2019bert}) are not capable of sequence generation. 
This paper focuses on PLMs having both encoder blocks and decoder blocks, such as T5~\cite{colin2020t5} and BART~\cite{lewis2020bart}.
% In fact, both T5 and BART are complete Transformer structures with minor modifications. 
% such as T5 uses relative position embeddings.
% Please refer to their origial paper for more details.

An encoder-decoder PLM takes a sequence $\rvs = \left(s_1,..., s_n \right)$ as input, and outputs a sequence $\rmP\rmL\rmM(\rvs) = \rvo = \left(o_1,..., o_m \right)$.
Each PLM has an associated vocabulary $\sV$, which is a set of tokens that can be directly accepted and embedded by the PLM.
The PLM first splits the input sequence into tokens in the vocabulary with a subword tokenization algorithm, such as SentencePieces~\cite{kudo2018sentencepiece}.
Then, the tokens are mapped into vectors by looking up the embedding table.
The attention blocks digest the embedded sequence and generate the output sequence.
% The embedded sequence passes though the attention blocks and the decoder generates the output sequence.

% Words outside the vocabulary are split into smaller tokens in the vocabulary.

% Then we give the formal definition of Generation Pre-trained Language Model:

% \vpara{Input:} x = $\{t_1, t_2,..., t_n\}$ is the input sequence, where $t_i$ is the $i$-th token (character, sub-word or word).

% \vpara{Output:} y = $\{t^{'}_1, t^{'}_2,..., t^{'}_m\}$ is the output sequence, the meaning of $y$ is depend on specific task setting.

% \model~uses PLM with additional special tokens that are not in the vocabulary. We denote these tokens by wrapping them with square brackets.

% \begin{figure*}[t]
% \centering
%     \includegraphics[width=1.0\linewidth]{figures/fig2-modelv2-placehold.pdf}
% \caption{
% Model Figure Placehold
% }
% \label{fig:model}
% \end{figure*}

\section{Method}
\label{sec:method}
% In this section, we will describe the overall frame work of our methodology, including the generation PLMs and parsing serialization.

% \model~consists of a Serializer and a Positional Prompt.
% The Serializer converts the dependency structure into serialized representation, and vice versa.
% The Positional Prompt injects positional information to the input sentence.
% Figure~\ref{fig:model} shows the overall framework of \model.
% The following section will specifies these two modules and the model training method.

\model~leverages a PLM to parse the dependency relation of a sentence by sequence generation.
Therefore, the Serializer converts the dependency structure into a serialized representation that meets the output format of the PLM (Section~\ref{sec:serializer}).
The Positional Prompt injects word position information into the input sentence so as to avoid numerical reasoning (Section~\ref{sec:pp}).
The PLM is modified by adding special tokens introduced by the Serializer and the Positional Prompt (Section~\ref{sec:plm}).
Figure~\ref{fig:model} illustrates the overall framework.

\subsection{Serializer for Dependency Structure}
\label{sec:serializer}

The Serializer $\rmS: (\rvx, \rvy) \mapsto \rvt$ is a function that maps sentence $\rvx$ and its corresponding dependency pairs $\rvy$ into a serialized representation $\rvt$, which servers as the target output to fine-tune the language model.
The Inverse Serializer $\rmS^{-1}: (\rvx, \rvo) \mapsto \rvy$ converts the output $\rvo$ of the PLM into dependency pairs to meet the output requirement of the DP task.
% dependency parsing task.

Specifically, the Serializer $\rmS$ decomposes dependency pairs, $\left\{\left(h_i^j, r_i^j\right)\right\} \in \rvy$, into smaller dependency units by scattering the dependent word $w_i$ into each of its head word, which forms the following triplets set: $\left\{ (w_i, r_i^j, h_i^j) \right\}$.
Then, it replaces each relation $r_i^j$ with a special token\footnote{Brackets indicate special tokens out of vocabulary $\sV$.}
$\left[\mathtt{REL}\left(r_i^j\right)\right] \in \sR$, where $\sR$ is a set of special tokens for all different relations.
The head word $h_i^j$ is substituted by its position in the input sentence $\rvx$, denoted as $\mathtt{POS}\left(h_i^j\right)$.
The target serialized representation $\rvt = \rmS(\rvx)$ concatenates all the dependency units with split token $[\mathtt{SPT}]$ as the following:
\begin{equation*}
\Big(...\Big[\mathtt{SPT}\Big]\ \underbrace{w_i\ \Big[\mathtt{REL}\left(r_i^j\right)\Big]\ \mathtt{POS}\left(h_i^j\right)}_{\text{\small one dependency unit}}\ \Big[\mathtt{SPT}\Big]...\Big)
\end{equation*}
The Inverse Serializer $\rmS^{-1}$ restores the dependency structure from the serialized representation by substituting the special token $\left[\mathtt{REL}\left(r_i^j\right)\right]$ with the original relation and indexing the head with its position $\mathtt{POS}\left(h_i^j\right)$ in the input sentence $\rvx$.

There are two issues in the Serializer designing:

\textbf{Word Ambiguity.}
It is highly possible to have words, especially function words, appear multiple times in one sentence, e.g., there are more than $72\%$ sentences in Penn Treebank~\cite{marcus1993ptb3} have repeated words.
We take two measures for word disambiguation in a dependency unit:
(1) To disambiguate head word, the Serializer represents the head word by its position, rather than the word itself;
(2) To disambiguate dependent word, the Serializer arranges dependency units by order of the dependent word in the input sentence $\rvx$, rather than topological ordering or depth/breadth first search ordering of the dependency graph.
% The Inverse Serializer scans $\rvx$ and $\rvo$ simultaneously so as to refer each dependent word to $\rvx$.
% The Inverse Serializer scans $\rvx$ and $\rvo$ simultaneously so as to refer suitable dependent word to $\rvx$.
The Inverse Serializer scans $\rvx$ and $\rvo$ simultaneously so as to refer the corresponding dependent word to $\rvx$.

% \textbf{Isolated Words.}~There are dependency schemata allowing for isolated words which have neither head words nor dependency relations with other words, e.g., the period mark in the SeDP results shown in Figure~\ref{fig:parsingtree}.
\vpara{Isolated Words.}
There are dependency schemata allowing for isolated words which have neither head words nor dependency relations with other words, e.g., the period mark in the SeDP results shown in Figure~\ref{fig:parsingtree}.
Note that the isolated words are different from the root word, as the root word is the head word of itself.
% \textbf{\color{red}}
One direct solution is to remove the isolated words from the serialized representation.
However, this will result in inconsistencies between $\rvx$ and $\rvt$, which complicates the word disambiguation.
Thus, We use special token $[\mathtt{NO}]$ to denote such isolation relation and word $no$ to represent the position of the virtual head word.
% The dependency unit for isolated word $w$ is defined as:
% \begin{equation*}
% \left(w, [\mathtt{REL-NO}], [\mathtt{POS-NO}]\right)
% \end{equation*}

% \textbf{Dependency Unit.}
% Another option to construct dependency unit is to merge our dependen

% The dependency units are arranged by the word order in $\rvx$ for word disambiguation.
% The same word, especially function words, could appear more than ones in a sentence.

% Another reason is for word disambiguation in the Inverse Serializer.
% Words, especially function words could appear multiple times in a sentence.
% Inverse Serializer scan the model output and the input sentence simultaneously 

% This also eases the designing of the Inverse Serializer $\rmS^{-1}$ because the same word may appear multiple times in a sentence, especially for the function words.
% Keeping the 

% disambiguation

% which restore the dependency structure by scanning the output sequence $\rvo$ and find the head word of each word by indexing $\mathtt{POS}\left(h_i^j\right)$ in the input sentence $\rvx$.

% There are other possible methods, such as $w_i, rel_1, pos_1, rel_2, pos_2$, but it results in inconsistent units, complicates the learning process.

% How to tackle no head word

% The word outside dependency graph is attached special token $[no]no$ to represent it has no head word and dependency relation with other words.

\subsection{Positional Prompt for Input Sentence}
\label{sec:pp}
As Section~\ref{sec:serializer} mentions, representing the head words by their positions is an important scheme for head word disambiguation.
However, PLMs are less skilled at numerical reasoning~\cite{geva2020injecting}.
We also empirically find it difficult for the PLM to learn the positional information of each word from scratch.
Thus, we inject Positional Prompt (PP) for each word, which converts the positional encoding problem into generating the position number in the input, rather than counting for each word.

In particular, given the input sentence $\rvx$, the positional prompt is the position number of each word $w_i$ wrapped with two special tokens $[\mathtt{PID}]$ and $[\mathtt{SPT}]$.
$[\mathtt{PID}]$ marks the beginning of the position number and prevents the tokenization algorithms from falsely taking the position prompt as part of the previous word.
% We do not use space character because space indicates whether a word is the first word of a sentence.
% We do not use space character to avoid potential
$[\mathtt{SPT}]$ separates the position number from the next word.
They also provide word segmentation information for some languages, such as Chinese.
After the conversion, we have the input sequence in the following form:
\begin{equation*}
    \rvs = w_1\ [\mathtt{PID}]\ 1\ [\mathtt{SPT}]\ w_2\ [\mathtt{PID}]\ 2\ [\mathtt{SPT}] \cdots
\end{equation*}
For brevity, we denote the above process as a function $\rmP\rmP: \rvx \mapsto \rvs$ that maps input sentence into sequence with positional prompt.

% \subsection{PLM Preparation (Implementation Details)}
% \subsection{Preparing Special Tokens}
\subsection{PLM for Sequence Generation}
\label{sec:plm}

Both Serializer and Positional Prompt introduce special tokens that are out of the original vocabulary $\sV$, including the relation tokens in $\sR$, the separation tokens $[\mathtt{PID}], [\mathtt{SPT}]$, and the special relation token $[\mathtt{NO}]$.
Before training, these tokens are added to the vocabulary, and their corresponding embeddings are randomly initialized from the same distribution as other tokens.
% function information learn from training
As we should notice, these special tokens are expected to undertake different semantics.
PLM thus treats them as trainable variables and learns their semantics during training.

% To avoid potential tokenization error, we augment the vocabulary of the PLM with these special tokens in the preparation stage.
% The input embedding of the special tokens are randomly initialized from the same distribution as other tokens.

% \subsection{Inference and Training}

% \textbf{Inference Data Flow.}
With all the three components of \model, input sentence is first converted into sequence with positional prompt: 
$\rvs = \rmP\rmP(\rvx)$.
The sequence is further fed into the PLM and get the sequence output with the maximum probability: 
$\rvo = \rmP\rmL\rmM(\rvs)$.
The final predicted dependency structure is recovered via the Inverse Serializer:
$ \rvy' = \rmS^{-1}(\rvo) $.

% \textbf{Training Objective.}
The training objective aims to maximize the likelihood of the ground truth dependency structure.
To do so, we take the serialized dependency structure as the target and minimize the auto-regressive language model loss.
% \textbf{Intermediate Fine-tuning.}
We can further enhance the unsupervised cross-domain capacity of \model~with intermediate fine-tuning (IFT)~\cite{pruksachatkun2020intermediate,chang2021rethinking}.
Before training on the dependency parsing, the intermediate fine-tuning uses the unlabeled sentences in the target domain and continues to train the PLM in source domain.

\section{Experiments}

\subsection{Evaluation Setups}

\subsubsection{Datasets}
We evaluate DPSG on the following $4$ widely used benchmarks for both SyDP and SeDP. We show more details about datasets in Appendix~\ref{app:data}.

\begin{itemize}[leftmargin=*]
\setlength\parskip{0pt}
\setlength\itemsep{3pt}
    \item \textbf{Penn Treebank} (PTB)~\cite{marcus1993ptb3} is the most proverbial benchmark for SyDP.
    \item \textbf{Chinese Open Dependency Treebank} (CODT) \cite{li2019codt} aims to evaluate the cross-domain SyDP capacity of the parser.
    It includes a balanced corpus (BC) for training, and three other corpora gathering from different domains for testing: product blogs (PB), popular novel ``Zhu Xian'' (ZX), and product comments (PC).
    \item \textbf{BroadCoverage Semantic Dependency Parsing} dataset (SDP15)~\cite{stephan2015sdp15} annotates English SeDP sentences with three different schemata, named as DM, PAS, and PSD.
    % and Chinese sentences with PSD, Czech setnences with PSD.
    It provides both in-domain (ID) and out-of-domain (OOD) evaluation datasets.
    The schema of SDP15 allows for isolated words.
    % in the dependency graph.
    % SDP15 adopts the schema that allowing isolated words in the dependency semantic graph.
    \item \textbf{Chinese semantic Dependency Parsing} dataset (SDP16)~\cite{che2016semeval} is a Chinese SeDP benchmark.
    The sentences are gathered from News (NEWS) and textbook (TEXT).
    The schema of SemEval16  allows for multiple head words but does not have isolated words.
\end{itemize}

\subsubsection{Evaluation Metrics}
% We follow the convention in DP to set evaluation metrics.
Following the conventions, we use unlabeled attachment score (UAS) and labeled attachment score (LAS) for SyDP.
We use labeled attachment F1 Score (LF) on SDP15 of SeDP.
For SeDP on SemEval16, we use unlabeled attachment F1 (UF) and labeled attachment F1 (LF).
All the results are presented in percentages ($\%$).
 
% The metrics used for SyDP are unlabeled attachment score (UAS) and labeled attachment score (LAS). 
% As there are non-determinant number of head words in SeDP, We evaluate SeDP results with labeled attachment F1 Score (LF).
% Considering the number of arcs is not determined, labeled attachment F1 Score (LF) is used for SeDP.

\subsubsection{Implementations} 
We use T5-base~\cite{colin2020t5} and mT5-base~\cite{xue2021mt5} as the backbone PLM for English dependency parsing and Chinese dependency parsing, respectively.
In particular, we use their V1.1 checkpoints, which are only pre-trained on unlabeled sentences, so as to keep the PLM unbiased.
In order to focus on the parsing capability of PLM itself, we do not use additional information, such as part-of-speech~(pos) tagging and character embedding~\cite{wang2020second,gan21mrc}.

The PLM is implemented with Huggingface Transformers~\cite{wolf2020transformers}. The learning rate is $4\times e^{-5}$, weight decay is $1\times e^{-5}$. 
The optimizer is AdamW~\cite{loshchilov2017adamw}. We conduct all the experiments on Tesla V100. 
% \zijun{We can put this paragraph in the appendix.}

\subsection{Baselines}

% As previous methods are difficult to perform generally well on both SyDP and SeDP when evaluating on both in-domain and cross-domain corpora.
% under both in-domain and cross-domain setting.
We divide baselines into three main categories based on their domain of expertise. 
% It should be noted
Note that almost all baselines use the additional lexical-level feature (including pos tagging, character-level embedding, and other pre-trained word embeddings), which is different from our \model.
We supplement more details about baselines in Appendix~\ref{app:baseline}.

\textbf{In-domain SyDP.}
\textit{Biaffine}~\cite{dozat2017biaffine}, \textit{StackPTR}~\cite{ma2018stackptr}, and \textit{CRF2O}~\cite{li2020crf} introduce specially designed parsing modules without PLM.
\textit{CVT}~\cite{clark2018cvt}, \textit{MP2O}~\cite{wang2020second}, and \textit{MRC}~\cite{gan21mrc} are recently proposed PLM-based dependency parser.
\textit{SeqNMT}~\cite{li2018seq2seq}, \textit{SeqViable}~\cite{strzyz2019viable}, and \textit{PaT}~\cite{vacareanu2020pat} cast dependency parsing as sequence labeling task, which is closely related to our sequence generation method.
% No part-of-speech information is used in all the sequence-based results (including DPSG).

\textbf{Unsupervised Cross-domain SyDP.}
\citet{peng2019nlpcc} and \citet{li2019codt} modify the \textit{Biaffine} for the unsupervised cross-domain DP.
\textit{SSADP}~\cite{lin2021unsupervised} relies on extra domain adaptation steps.
In the PLM era, \citet{li2019codt} propose \textit{ELMo-Biaffine} with IFT on unlabeled target domain data.
% \citet{li2019bert} intermediate fine-tune the BERT on unlabeled target domain data and use tri-training to ensemble three fine-tuned BERT parsers.
% \zijun{add bert}.

\textbf{SeDP.}
\citet{dozat2018sedp} modify  \textit{Biaffine} for SeDP.
\textit{BS-IT}~\cite{wang2019sedp} is a transition-based semantic dependency parser with incremental Tree-LSTM.
\textit{HIT-SCIR}~\cite{che2019pipeline} solves the SeDP with a BERT based ipeline. \textit{BERT+Flair}\footnote{They use different pre-processing scripts on SDP15, thus are not comparable with \model~and other baselines on SDP15.}~\cite{he2020bert} augments the Biaffine model with BERT and Flair~\cite{akbik2018flair} embedding.
\textit{Pointer}~\cite{fernandez2020transition} combines transition-based parser with Pointer Network. 
It is also augmented with a Convolutional Neural Network (CNN) encoder for the character-level feature.
% they cast the original SDP format into the CONLL-X format by adding the isolated words into the DAG.

% The most different setting between our DPSG and other baselines is our model do not utilize additional lexical-level features~(part-of-speech tagging, character-embedding and other pre-trained word embeddings). Our DPSG encodes feature only from the token itself, which can show the parser potential of PLM clearly.

\subsection{Main Results}
% Table~\ref{tab:exp:ptb3:result}, Table~\ref{tab:sem16:result}, Table~\ref{tab:exp:sdp15} and Table~\ref{tab:exp:codt1:result} list the results of our \model~and baselines.

\subsubsection{\model~is Schema-Free}
% All the results demonstrate our DPSG is a schema-free method, which mainly reflected in the following three points.
% The schema-free characteristics of \model~ are reflected by the following three perspectives.
The schema-free characteristics of \model~ are reflected by the following two perspectives.

\begin{table}[t]
\centering
\scalebox{0.92}{
    \setlength{\tabcolsep}{3pt}
    \begin{tabular}{clcc}
    \toprule
    % \multirow{2}{*}{} & \multicolumn{2}{l}{\makecell[c]{PTB3}} \\
        %  & UAS & LAS \\
    Features & Method (PLM) & UAS & LAS \\
    \midrule
    Char & CRF2O & $96.14$ & $94.49$ \\
    POS & Biaffine & $95.74$ & $94.08$ \\
    POS & StackPTR & $95.87$ & $94.19$ \\
    % \multicolumn{3}{l}{\makecell[c]{without PTM}} \\
    % Biaffine & 95.74 & 94.08 \\
    % StackPTR & 95.87 & 94.19 \\
    % BiaffinGNN & 95.87 & 94.15 \\
    % BiaffineCRF & 96.14 & 94.49 \\
    % \midrule
    % \multicolumn{3}{l}{\makecell[c]{with PTM}} \\
    Char+POS & $^{\dag}$MP2O (BERT{\small-large})& $96.91$ & $95.34$ \\
    POS & $^{\dag}$MRC (RoBERTa{\small-large}) & $\mathbf{97.24}$ & $\mathbf{95.49}$ \\
    % \midrule
    % \multicolumn{3}{l}{\makecell[c]{Seq2Seq}} \\
    POS & $^{\dag}$CVT (CVT) & $96.60$ & $95.00$ \\
    \midrule
    POS & $^{\ddag}$SeqNMT & $92.08$ & $94.11$ \\
    POS & $^{\ddag}$SeqViable & $93.67$ & $91.72$ \\
    POS & $^{\dag\ddag}$PaT (BERT{\small-base}) & $95.87$ & $94.66$ \\
    % \cmidrule(2-4)
    \cmidrule(lr){1-4}
    - & $^{\dag\ddag}$\model~(T5{\small-base}) & $\mathbf{96.48}$ & $\mathbf{95.04}$ \\
    - & $^{\dag\ddag}$\model~(Multi) & $96.25$ & $94.85$ \\
    \bottomrule
    \end{tabular}
}
\caption{
Results on PTB for SyDP. 
Features means these methods use additional lexical-level information, such as character embedding (Char) or part of speech tagging (POS).
${\ddag}$ means this method belongs to sequence based methods.
${\dag}$ means this method use PLM, and the used PLM as listed in parenthesis.
}
\label{tab:exp:ptb3:result}
\end{table}

% \begin{table}[h]
% \centering
% \scalebox{1.0}{
%     \begin{tabular}{llrr}
%     \toprule
%     % \multirow{2}{*}{} & \multicolumn{2}{l}{\makecell[c]{PTB3}} \\
%         %  & UAS & LAS \\
%     Category & Method & UAS & LAS \\
%     \midrule
%     \multirow{3}{*}{\rotatebox{0}{w/o PLM}}
%     & Biaffine &  95.74 & 94.08 \\
%     & StackPTR & 95.87 & 94.19 \\
%     % & BiaffinGNN & 95.87 & 94.15 \\
%     & CRF2O & 96.14 & 94.49 \\
%     % \multicolumn{3}{l}{\makecell[c]{without PTM}} \\
%     % Biaffine & 95.74 & 94.08 \\
%     % StackPTR & 95.87 & 94.19 \\
%     % BiaffinGNN & 95.87 & 94.15 \\
%     % BiaffineCRF & 96.14 & 94.49 \\
%     \midrule
%     % \multicolumn{3}{l}{\makecell[c]{with PTM}} \\
%     \multirow{3}{*}{\rotatebox{0}{with PLM}} 
%     & CVT & 96.60 & 95.00 \\
%     & MP2O & 96.91 & 95.34 \\
%     & MRC & 97.24 & 95.49 \\
%     \midrule
%     \multirow{4}{*}{\rotatebox{0}{Seq2Seq}} 
%     % \multicolumn{3}{l}{\makecell[c]{Seq2Seq}} \\
%     & SeqNMT & 92.08 & 94.11 \\
%     & SeqViable & 93.67 & 91.72 \\
%     & PaT & 95.87 & 94.66 \\
%     % \cmidrule(2-4)
%     \cmidrule(lr){2-4}
%     & \model & 96.48 & 95.04 \\
%     \bottomrule
%     \end{tabular}
% }
% \caption{Results on PTB3 for SyDP}
% \label{tab:exp:ptb3:result}
% \end{table}
\begin{table}[t]
\centering
\scalebox{0.92}{
\setlength{\tabcolsep}{7pt} % Default value: 6pt
\begin{tabular}{lcccccc}
    \toprule
    \multirow{2}{*}{Method}
    & \multicolumn{2}{c}{NEWS} & \multicolumn{2}{c}{TEXT} \\
    \cmidrule(lr){2-3}\cmidrule(lr){4-5}
    & \begin{tabular}{c} UF \end{tabular}
    & \begin{tabular}{c} LF \end{tabular}
    & \begin{tabular}{c} UF \end{tabular}
    & \begin{tabular}{c} LF \end{tabular} \\
    \midrule
    % Dounar & 77.64 & 59.06 & 82.41 & 68.59 \\
    BS-IT & $81.14$ & $63.30$ & $85.71$ & $72.92$ \\
    BERT+Flair & $82.92$ & $67.27$ & $91.10$ & $80.41$ \\
    \midrule
    \model & $\textbf{84.31}$ & $\textbf{70.82}$ & $\textbf{90.97}$ & $\textbf{82.36}$ \\
    \bottomrule
\end{tabular}
}
\caption{Experimental results on SemEval16. }
\label{tab:sem16:result}
\end{table}
\begin{table}[t]
\centering
\scalebox{0.9}{
\setlength{\tabcolsep}{6pt} % Default value: 6pt
\begin{tabular}{lcccccccc}
\toprule
% \multirow{2}{*}{Method} 
% & \multicolumn{2}{c}{DM} & \multicolumn{2}{c}{PAS} 
% & \multicolumn{2}{c}{PSD} & \multicolumn{2}{c}{AVG} \\
% \cmidrule(lr){2-3}\cmidrule(lr){4-5}\cmidrule(lr){6-7}\cmidrule(lr){8-9}
% & ID & OOD & ID & OOD & ID & OOD & ID & OOD \\
Method (ID) & DM & PAS & PSD\\
\midrule
BS-IT & $90.3{\color{white}0}$ & $91.7{\color{white}0}$ & $78.6{\color{white}0}$ \\
% Peng & 90.4 & 92.7 & 78.5 & 87.2 \\
Biaffine & $93.7{\color{white}0}$ & $93.9{\color{white}0}$ & $81.0{\color{white}0}$ \\
$^{\dag}$HIT-SCIR~(BERT{\small-base}) & $92.9{\color{white}0}$ & $94.4{\color{white}0}$ & $81.6{\color{white}0}$ \\
% BERT+Flair & $\textbf{94.57}$ & $\textbf{96.13}$ & $\textbf{86.80}$ \\
$^{\dag}$Pointer~(BERT{\small-base}) & $94.4{\color{white}0}$ & $95.1{\color{white}0}$ & $\textbf{82.6{\color{white}0}}$ \\
\midrule
$^{\dag}$\model & $93.96$ & $94.26$ & $81.98$ \\
% \model~(PTB) & $94.24$ & $94.42$ & $82.11$ \\
$^{\dag}$\model~(Multi) & $\textbf{94.45}$ & $\textbf{96.59}$ & $82.25$ \\
\bottomrule
\toprule
Method (OOD) & DM & PAS & PSD\\
\midrule
BS-IT & $84.9{\color{white}0}$ & $87.6{\color{white}0}$ & $75.9{\color{white}0}$ \\
% Peng & 85.3 & 89.0 & 76.4 & 83.6 \\
Biaffine & $88.9{\color{white}0}$ & $90.6{\color{white}0}$ & $79.4{\color{white}0}$ \\
$^{\dag}$HIT-SCIR~(BERT{\small-base}) & $89.2{\color{white}0}$ & $92.4{\color{white}0}$ & $81.0{\color{white}0}$ \\
% BERT+Flair & $90.86$ & $\textbf{94.38}$ & $79.48$ \\
$^{\dag}$Pointer~(BERT{\small-base}) & $\textbf{91.0{\color{white}0}}$ & $\textbf{93.4{\color{white}0}}$ & $\textbf{82.0{\color{white}0}}$ \\
\midrule
$^{\dag}$\model & $90.47$ & $92.38$ & $80.04$ \\
% \model~(PTB) & $ 90.90$ & $92.40$ & $79.87$ \\
$^{\dag}$\model~(Multi) & $90.70$ & $92.31$ & $79.65$ \\
\bottomrule
\end{tabular}
}
\caption{
Experimental results on SDP15 in terms of LF.
\model~(Multi) means the parameters are optimized in the combination of PTB and current SeDP dataset.
$^{\dag}$ means the model utilizing PLM.
}
\label{tab:exp:sdp15}
\end{table}

\textbf{Towards Specific Schema.}
% Comparing with previous baselines, 
\model~obtains the SOTA performance on both CODT in Table~\ref{tab:exp:codt1:result} and SemEval16 in Table~\ref{tab:sem16:result}, and achieves the first-tier even among methods used additional lexical-level features on PTB in Table~\ref{tab:exp:ptb3:result} and SDP15 in Table~\ref{tab:exp:sdp15}.
% \boda{This is the general summary, next will add specific describes, we can solve parsing in this schema, that schema......}
For in-domain SyDP in Table~\ref{tab:exp:ptb3:result}, \model~outperforms all the previous sequence-based methods,
% on PTB w.r.t., UAS and LAS
and performs sightly lower than MRC, which uses contextual interactive pos tagging, by 0.45\% in LAS.

For SeDP in Table~\ref{tab:sem16:result}, \model~ourperforms BERT +Flair to a large margin on SemEval16, achieves 3.55\% performances gain on NEWS, and 1.95\% performance gain on TEXT with regard to LF.
\model~also outperforms the PLM-based pipeline HIT-SCIR on SDP15~(Table~\ref{tab:exp:sdp15}), but sightly lower than Pointer, which applies additional CNN to encode the character-level embeddings.
We also observe that \model~and the Pointer have the largest gap in the PSD schema of SDP15.
This is caused in that PSD has much more relation labels than the other schemata~\cite{peng2017multi}, which increases the search space of our generation model.

% \vpara{Cross-Schema.}
% We train the \model~on PTB with 20 epochs, and use the checkpoint to continue training in three schemata from SDP15.
% The results are shown on Table~\ref{tab:exp:ptb3:result}.
% We can observe the slight improvement in all test sets except the OOD-test of PSD. 
% This phenomenon proves our \model~is capable of the cross-schema parsing and also expose that syntactic dependency relation may be helpful for the training of semantic dependency parsing.

% the results are not the final results

\begin{table*}[t]
    \centering
    \scalebox{0.9}{
        \setlength{\tabcolsep}{6pt} % Default value: 6pt
        \begin{tabular}{llccccccccccccc}
        \toprule
        \multirow{2}{*}{Category} & \multirow{2}{*}{Model} & \multicolumn{2}{l}{\makecell[c]{BC$\rightarrow$ PB}}
         & \multicolumn{2}{l}{\makecell[c]{BC$\rightarrow$ ZX}}
         & \multicolumn{2}{l}{\makecell[c]{BC$\rightarrow$ PC}}
         & \multicolumn{2}{l}{\makecell[c]{Average}} \\
          \cmidrule(lr){3-4}\cmidrule(lr){5-6}\cmidrule(lr){7-8}\cmidrule(lr){9-10}
        & & UAS & LAS & UAS & LAS & UAS & LAS & \makecell[c]{UAS} & \makecell[c]{LAS} \\
        \midrule
        \multirow{2}{*}{w/o PLM} & Biaffine & $67.75$ & $60.95$ & $69.41$ & $61.55$ & $39.95$ & $26.96$ & $59.04$ & $49.82$\\
        & SSADP & $68.55$ & $61.59$ & $70.82$ & $63.61$ & $41.10$ & $27.67$ & $60.16$ & $50.96$ \\
        \midrule
        % ELMo & 70.91 & 64.70 & - & - & 44.31 & 30.46 \\
        \multirow{3}{*}{w/ PLM} & ELMo-Biaffine w/ IFT & $77.15$ & $71.54$ & $74.68$ & $67.51$ & $53.04$ & $39.48$ & $68.29$ & $59.51$\\
        % & BERT w/ IFT + ensemble & 81.61 & 76.77 & 79.74 & 74.32 & 60.50 & 49.49 & 73.95 & 66.86 \\
        \cmidrule(lr){2-10}
        & \model~w/o IFT & $78.86$ & $73.28$ & $75.74$ & $69.42$ & $54.00$ & $41.98$ & $69.53$ & $61.56$\\
        &\model~w/ IFT & $\mathbf{81.74}$ & $\mathbf{76.60}$ & $\mathbf{80.73}$ & $\mathbf{74.77}$ & $\mathbf{62.44}$ & $\mathbf{49.97}$ & $\mathbf{74.97}$ & $\mathbf{67.11}$ \\
        \bottomrule
        \end{tabular}
    }
    \caption{Results on CODT for unsupervised cross-domain SyDP.}
    \label{tab:exp:codt1:result}
\end{table*}

\textbf{Towards Multi-Schemata.}
Furthermore, we design the multi-schemata experiment.
We mix PTB and SDP15 by concatenating a prefix to the input text to distinguish different schemata.
% The train set of PTB is filter because some sentence are appeared in the test sets of SDP15.
To prevent data leakage, we filter out sentences from the training set of PTB, which also appear in the test set of SDP15. 
% This may be the reason of \model~(Multi) worse than \model~in Table~\ref{tab:exp:ptb3:result}.
As \model~(Multi) uses less training data for PTB, it performs worse than \model~in Table~\ref{tab:exp:ptb3:result}.
% The results are denoted as \model~(Multi) in Table~\ref{tab:exp:ptb3:result} and Table~\ref{tab:exp:sdp15}.
\model~(Multi) in Table~\ref{tab:exp:sdp15} outperforms Pointer by 1.49\% in ID evaluation of the PAS schema, 0.05\% in ID evaluation of the DM schema, and achieves almost the same performance with Pointer in ID evaluation of the PSD schema.
% There are significant improvement in ID of PAS.
The improvement over schema-specific model is most obvious on PAS.
It could be because the PAS schema is more similar to the syntax schema~\cite{peng2017multi}, thus it benefits more from PTB.
This multi-schemata approach also provides a new method to explore the inner connection between SyDP and SeDP.

\subsubsection{Unsupervised Cross-domain}
Table~\ref{tab:exp:codt1:result} demonstrates the outstanding transferability of \model.
% \boda{copy the specific descriptions from original version}
We implement DPSG with and without IFT on the target domain.
DPSG with IFT achieves the new SOTA, with a boosting of $5.06\%$, $7.21\%$ and $10.49\%$ in terms of LAS on PB, ZX, and PC, compared to ELMo with IFT.
% \boda{change the representation}
% One possible reason of the great performance of our \model~may because 
\model~is completely trained during IFT. While the additional biaffine module of ELMo cannot benefit from the unlabeled sentences from the target domain.

\section{Analysis}

This section studies whether there is better implementation for \model.
We are particularly interested in:
1) the designing of the Serializer, 
2) the effect of the introduced special tokens, 
and 3) the choice of the PLM model.
We use PTB as the benchmark and compare \model~introduced in Section~\ref{sec:method} with many other possible choices.
The results of these exploratory experiments are shown in Table~\ref{tab:exp:pilot:result}.
% We also provide analysis for the comparison results.

% In this section, we present some pilot experiments to empirically explore the designing of serialization and model choosing. The PLM used in model choosing experiment is BART-base, remaining experiments use T5-base. The datasets used for pilot experiments is PTB.

\subsection{Serializer Designing}
\begin{figure}[t]
\centering
    \includegraphics[width=0.98\linewidth]{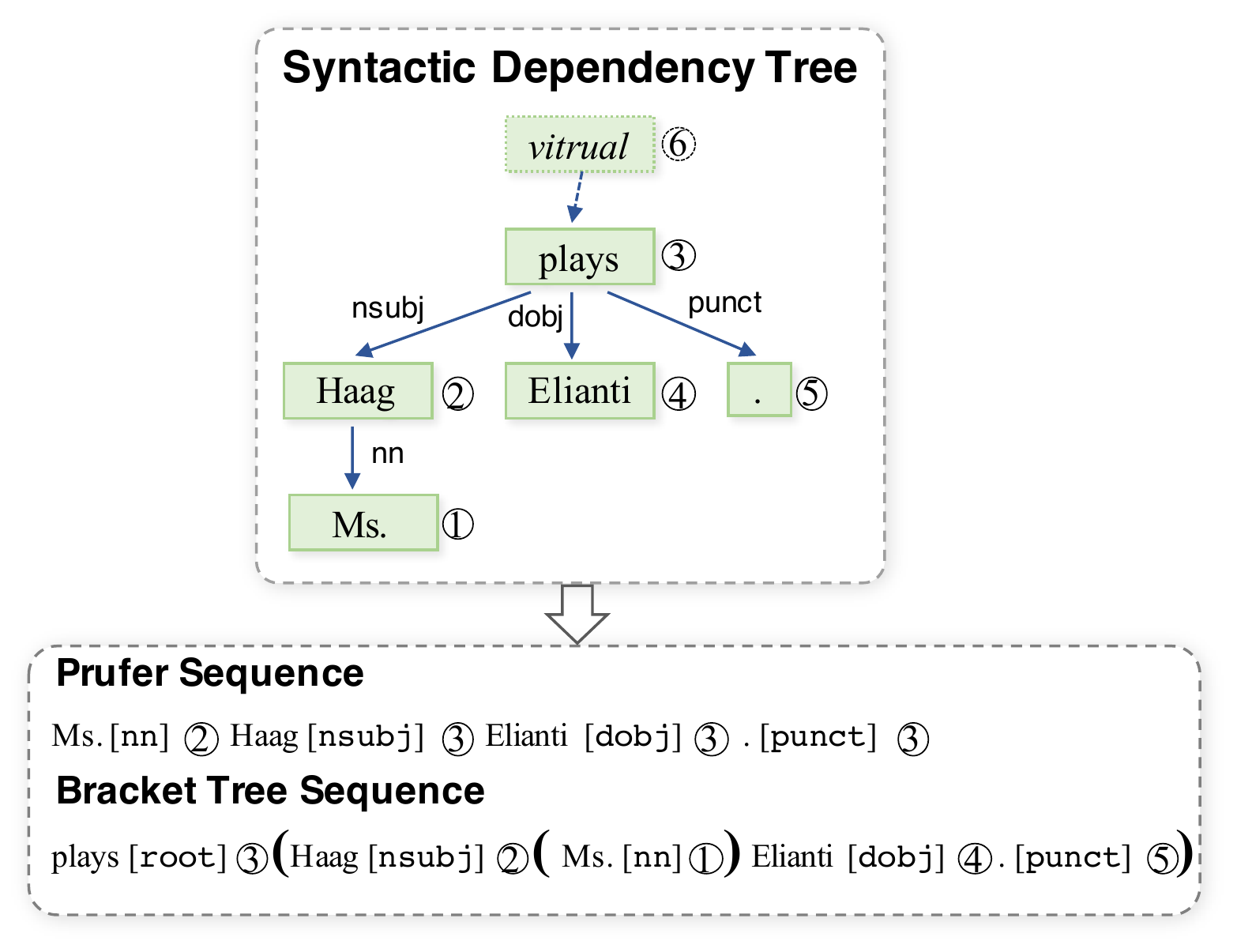}
\vspace{-0.1in}
\caption{
Prufer sequence and Bracket Tree sequence of the same sentence ``\textit{Ms. Haag plays Elianti .}''.
}
\label{fig:prufer-bracket}
\end{figure}

% Tree structure is a well-studied data structure, there are some other serialization methods converting tree structure into linear sequence.
Tree, as the well-studied data structure for syntactic dependency parsing, has several other serialization methods to be converted into serialized representations.
We explore the serializer designing of the tree structure in \model~with two other widely used serialized representation---Prufer sequence and Bracket Tree, which are shown in Figure~\ref{fig:prufer-bracket}.
Note that both Prufer sequence and Bracket Tree face the same word ambiguity issues; we associate each word with a unique position number as well.
% The serialization method we used in DPSG is to represent dependency relation in dependency units, and concatenate all the dependency units by the order of the input sentence.
% % concatenating dependency units together. 
% There are some other serialization method that maps tree-based dependency structure into sequence that can be directly generated by PLM.

% tree structure into the sequence, we also design the experiments for these methods.

\textbf{Prufer Sequence} is a unique sequence associated with the labeled tree in combinatorial mathematics.
% The only difference between labeled tree in combinatorial mathematics and dependency parsing is that the parsing tree is rooted tree, which lead to the ambiguity for the re-converting from prufer sequence to parsing tree.
The algorithm which converts labeled tree into Prufer sequence does not preserve the root node, while in dependency parsing, the root is a unique word.
% The only difference between labeled tree in combinatorial mathematics and dependency parsing is that the parsing tree is a rooted tree, while prufer sequence is designed for unrooted trees.
% which conflicts the precondition that the prufer sequence requires an unrooted tree.
To bridge this inconsistency, we introduce an additionally added virtual node to the dependency tree to mark the root word.
% with the maximum ID for each parsing tree.

\textbf{Bracket Tree} is one of the most commonly used serialization methods to represent the tree structure~\cite{strzyz2019viable}.
By recursively putting the sub-tree nodes in a pair of brackets from left-to-right, bracket tree can build a bijection between parsing tree and bracket tree.
% The details about the construction of bracket tree is shown in Appendix~\ref{app:bracket}.
% The details about the construction of Purfer Sequence in shown in Appendix~\ref{app:prufer}.
More details about how to construct the Prufer sequence and the bracket tree are shown in Appendix~\ref{app:prufer}.
% \zijun{You may say there are many other parsing methods that also use Bracket Tree, and cite them}

\begin{table}[t]
\centering
\scalebox{0.95}{
    \setlength{\tabcolsep}{5pt}
    \begin{tabular}{lcccc}
    \toprule
    Metric & \model & Prufer & Bracket \\
    \midrule
    UAS & $96.48$ & $85.53_{\downarrow 10.95}$ & $95.37_{\downarrow 1.11}$ \\
    LAS & $95.04$ & $83.72_{\downarrow 11.32}$ & $93.76_{\downarrow 1.28}$ \\
    \bottomrule
    \toprule
    Metric & \model$_\text{-pos}$ & \model$_\text{-rel}$ & \model$_\text{BART}$ \\
    \midrule
    UAS & $95.20_{\downarrow 1.28}$ & $93.88_{\downarrow 2.60}$ & $86.35_{\downarrow 10.13}$\\
    LAS & $93.17_{\downarrow 1.87}$ & $92.46_{\downarrow 2.58}$ & $79.45_{\downarrow 15.59}$\\
    \bottomrule
    \end{tabular}
}
\caption{Results on PTB for exploratory experiment}
\label{tab:exp:pilot:result}
\end{table}

    % \begin{tabular}{lccc}
    % \toprule
    % Metric & \model & Prufer & Bracket\\
    % \midrule
    % UAS & $96.48$ & $-$ & $-$ \\
    % LAS & $95.04$ & $84.92$ & $92.71$ \\
    % \bottomrule
    % \end{tabular}

% \begin{table}[h]
% \centering
% \begin{tabular}{cccccc}
% \toprule
% Methods & UAS & LAS \\
% \midrule
% \model & 96.48 & 95.04 \\
% DPSG$_\text{-pos}$ & 95.20 & 93.17 \\
% DPSG$_\text{-rel}$ & 93.88 & 92.46 \\
% BART & \\
% \bottomrule
% \end{tabular}
% \caption{Pilot experiments on PTB3.}
% \label{tab:exp:pilot:result}
% \end{table}

% The comparison results about different serialization methods are shown , 
We denote the experimental results of Prufer sequence and bracket tree as Prufer and Bracket, respectively, in Table~\ref{tab:exp:pilot:result}.
Both Prufer sequence and bracket tree undermine the performance of \model~to a large margin, which indicates that our proposed Serializer provides a better serialized representation for the PLM to generate.
This is because our Serializer guarantees the dependency units in the output have the same order of the words in the input sentences, while Prufer sequence and bracket tree do not preserve the order.
Thus, our proposed \model~\textit{expands} the input sentence to generate the output sequence, while Prufer sequence and bracket tree based \model~\textit{reconstruct} the syntax dependency structure.
As expansion strategy has smaller generation space than reconstruction, the serialization representation proposed in Section~\ref{sec:serializer} eases the learning complexity of the PLM, and further brings better performance.

% \input{tables/exp.result.bracket}

% The serialization used in \model~does not change the word orders from input sentence.
% \model~is more like using a similar \textbf{\textit{expansion}} method to express syntax in the generated result.
% But Prufer Sequence and Bracket Tree Sequence are more like using similar \textbf{\textit{reconstruction}} methods to express syntax in the generated results because the word orders are changed in these two sequences.
% Although none of them retain the original word order, the Bracket Tree Sequence retains at least all the words in the input sequence, but the set of words in the Prufer Sequence is a subset of the input sequence.
% This means that the Prufer Sequence is more different from the input sequence.
% Expansion by orders is easier than reconstruction out of order, this also indicates that guiding the generated sequence to retain parts of the information in the input sequence as much as possible is helpful to reduce the learning difficulty of\model.

\subsection{Special Tokens Designing}

We further investigate whether the additionally introduced special tokens are useful.

\textbf{Relation Tokens.} There are two different ways to represent the dependency relations in the serialized representation: 
adding a special token for each dependency relation, or mapping each dependency relation to one token in the original vocabulary with the closest meaning, e.g., \textit{conj} $\to$ conjunct. 
Experimental results using word mapping is denoted as \model$_\text{-rel}$ in Table~\ref{tab:exp:pilot:result}.
% We conduct experiment for this situation, which is named as \model$_\text{-rel}$ in Table~\ref{tab:exp:pilot:result}.
\model$_\text{-rel}$ is inferior than \model, which indicates that the special tokens for relations are important.
The reason is that if we use the tokens in the original vocabulary, they interfere with their original meanings as the word.
Special tokens disentangle the dependency relation from the words that could appear in the sentence.

% one is adding the original dependency relations into the token sets of T5-tokenizer. The other is mapping the dependency relations into the complete words, e.g. \textit{conj}-> conjunct. We conduct experiment for this situation, which named is DPSG$_{-rel}$.
% The advantage of the first one is reducing are:  1) reducing the search space in decoding process 2) the relation token and special token may not in the original vocabulary set. We conduct experiments by mapping dependency relations from original label into the entire word. The name of this experiment is DPSG$_{-rel}$.

% \subsection{Position Prompt.}
% The representation of head word is one of the most important choosing in sequence. 

% \vpara{Word Representation.} Considering the DPSG is generate-based, directly generate head word is a neural chosen. But 

% \vpara{Split Token}

\textbf{Positional Prompt.} 
We are also particularly interested in the effectiveness of the positional prompts.
We conduct experiments where the positional prompt is removed and send the original input sentence to the PLM. 
The result is denoted as \model$_\text{-pos}$ in Table~\ref{tab:exp:pilot:result}.
\model$_\text{-pos}$ undermines the performance of \model~because it requires the PLM to perform numerical reasoning, that is, to count for the position of each head word.

% The positional prompt of the input sequence directly indicate the positions of the head words and reduce the learning difficulty of PLM. We also conduct experiment when dropping the position prompts from input sequences. The name of this experiment is DPSG$_\text{-pos}$.

% Comparing the performance among \model, \model$_{-pos}$ and \model$_{-rel}$, we can get two observations: 1) Decomposing complex problem (positioning problem and parsing problem) into simple problems (counting problem and parsing problem) is useful for \model. 2) Reducing the search space of generation can improve the performance for \model.

% \vpara{Single Prediction.}

\subsection{Model Choosing} 

Both BART and T5 are widely used encoder-decoder PLMs. 
We try BART-base as the backbone PLM in \model.
Table~\ref{tab:exp:pilot:result} shows that BART undermines the performance.
In addition, BART has a significant performance drop after achieving the best performance, as shown in Appendix~\ref{app:bart}. 

\subsection{Legality}
% There are two different legality in \model. One is the \textit{Formation Legality}, the other is \textit{Structure Legality}. \textit{Formation Legality} focus on whether the sequence having the legal representation, e.g. each \textit{dependency unit} should include three parts in the choosing serialization. While \textit{Structure Legality} focus on whether the sequence can represent the legal tree structure (no loop) or legal directed acylic graph.
% Considering there is not hard-constraint in the generation of \model, we count the legality of the SyDP sequence generated by DPSG. The statistics show \model~generate $4103/4116$ legal sentences of the inputs from dev and test sets of PTB. The legal ratio is $99.68\%$, which is acceptable.
There are two different legalities in \model. \textit{Formation Legality} focus on whether the sequence has the correct formation~(see Section~\ref{sec:serializer}) and \textit{Structural Legality} focus on the legality of the corresponding parsing structure.
The statistics on PTB show that the formation legality of \model~is $100\%$, and the structure legality of \model~is $99.7\%$, which is acceptable in practical usage.

% Considering the DPSG solve parsing only depends on PLM, it is worthy to explore the influence of the model choosing. We conduct experiments on T5-base and BART-base~\cite{lewis2020bart}. It is obvious that T5 outperforms BART. Beyond this, BART also has a significant performance drop after achieving the best performance, which is shown in Appendix~\ref{app:bart}. 

% \subsection{Observations}
% % From these results of pilot experiments, we can get some conslusions: 1) Reducing the learning problem 2) Reducing the search space 3) let each token overtake single semantic function is useful

% We have some observations from the results in Table~\ref{tab:exp:pilot:result}: 1) Reducing the search space of generation is helpful for DPSG. 2) Decomposing complex problem into simple problems is helpful for DPSG. 3) The simpler the representation of the sequence, the better the performance of the model. 4) The T5 model significantly outperform the BART.
\section{Related Work}

\subsection{Syntactic Dependency Parsing}
\textbf{In-domain SyDP.}
Transition-based methods and graph-based methods are widely used in SyDP.~\citet{dozat2017biaffine} introduce biaffine attention into the graph-based methods.~\citet{ma2018stackptr} adopt pointer network to alleviate the drawback of local information in transition-based methods.
% ~\citet{ji2019graph} employ GNN to capture high-order information in graph-based methods.
~\citet{li2020crf} improve the CRF to capture second-order information.

% The research of sequence-based methods also attract attention.
% There are also researches about achieving SyDP by sequence to sequence methods.
There are also researches using sequence to sequence methods for SyDP.
\citet{li2018seq2seq} use BiLSTM to predict the labeling of positions and relations of dependency parsing.~\citet{strzyz2019viable} improve~\citet{li2018seq2seq}'s method and explore more representation of predicated labeling sequence of dependency parsing.~\citet{vacareanu2020pat} use BERT to augment the sequence labeling methods.

% Comparing to these method, \citet{gan21mrc} adopt Machine Reading Comprehension (MRC) framework to achieve span-based dependency parsing and get the SOTA performance.

\textbf{Unsupervised Cross-domain SyDP.}
The labeling of parsing data requires a wealth of linguistics knowledge and this limitation facilitates the research of unsupervised cross-domain DP.~\citet{yu2015self} introduce pseudo-labeling unsupervised cross-domain SyDP via self-training.~\citet{li2019codt} propose a cross-domain datasets CODT for SyDP and build baselines for unsupervised cross-domain SyDP.~\citet{lin2021unsupervised} introduce feature-based domain adaptation method in this field.
% There are two types of cross-domain DP: semi-supervised cross-domain DP and unsupervised cross-domain DP. The unsupervised cross-domain DP is harder but more valuable.
% \citet{li2019bert} use BERT \cite{devlin2019bert} and model-ensemble for unsupervised SyDP.

\subsection{Semantic Dependency Parsing}
% \citet{li2003sdp} released the first large-scale Chinese semantic dependency corpus.
\citet{jan2017} accomplish the first transition-based parser for Minimal Recursion Semantics (MRS).
\citet{zhang2016} present two novel transition-systems to generate arbitrary directed graphs in an incremental manner.
\citet{dozat2018sedp} modify the Biaffine~\cite{dozat2017biaffine} for SeDP. 
However, due to the words in SeDP may have multiple-head, there is not sequence-based method for SeDP now.

\subsection{Probing in Language Model}
The research of exploring whether PLM can learn the linguistic features during the pre-training process, especially syntax knowledge, attracts some attention.~\citet{hewitt2019structural} map the distance between word embedding in PLM into the distance in syntax tree and construct a syntax tree without relation label.~\citet{clark2019bert} design a structural probe to detect the ability of attention heads to express \textit{dobj} (direct object) dependency relation.
Their results prove the syntax knowledge can also be found in the attention maps.
% The results show that BERT even can detect the more complex syntax relation such as \textit{coreference}.
% The results showed that the accuracy of \textit{dobj} was as high as $86.8\%$ in PTB.

% I/O
\section{Conclusion}
This paper proposes \model---a schema-free dependency parsing method.
By serializing the parsing structure to a flattened sequence, PLM can directly generate the parsing results in serialized representation.
\model~not only achieves good results in each different schema, but also performs surprisingly well on unsupervised cross-domain DP.
The multi-schemata experiments also suggest
that \model~is capable of investigating the inner connection between different schemata dependency parsing.
The exploratory experiments and analyses demonstrate the rationality of the designing of \model.
Considering the unity, indirectness, and effectiveness of \model, we believe it has the potential to become a new paradigm for dependency parsing.

% In this paper, we propose solve schema-free dependency parsing via sequence generation (DPSG).
% By serializing the parsing structure into sequence, the dependency parsing can be solved as the sequence generation problem.
% We explore different designing of serializing and the model choosing, the final choice shows good performance on multiple schema of dependency parsing.
% Comparing the previous parsing methods, our \model accomplish different schemata parsing depend on the same model without additional parsing decoder and lexical-level features.
% The another advantage of DPSG is the potential of achieve multi-schema and cross-schema parsing.
% This approach is also beneficial for exploring the inner relationship between Syntactic Dependency Parsing and Semantic Dependency Parsing.
% In addition, our research demonstrate PLM is capable a great parser itself.
% By replacing the serialization with Prufer Sequence and Bracket Tree Sequence, we observe the significantly performance drop in PTB.

% Despite the great performance of DPSG, there are still some future work are valuable to research.
% First is the 

% \input{section/8.lipsum}

\clearpage
\bibliographystyle{acl_natbib}
% \bibliography{1.ref}
\bibliography{ref_output}

\begin{thebibliography}{45}
\expandafter\ifx\csname natexlab\endcsname\relax\def\natexlab#1{#1}\fi

\bibitem[{Akbik et~al.(2018)Akbik, Blythe, and Vollgraf}]{akbik2018flair}
Alan Akbik, Duncan Blythe, and Roland Vollgraf. 2018.
\newblock \href {https://aclanthology.org/C18-1139} {Contextual string
  embeddings for sequence labeling}.
\newblock In \emph{Proceedings of the 27th International Conference on
  Computational Linguistics}, pages 1638--1649, Santa Fe, New Mexico, USA.
  Association for Computational Linguistics.

\bibitem[{Bugliarello and Okazaki(2020)}]{bugliarello2020mt}
Emanuele Bugliarello and Naoaki Okazaki. 2020.
\newblock \href {https://doi.org/10.18653/v1/2020.acl-main.147} {Enhancing
  machine translation with dependency-aware self-attention}.
\newblock In \emph{Proceedings of the 58th Annual Meeting of the Association
  for Computational Linguistics}, pages 1618--1627, Online. Association for
  Computational Linguistics.

\bibitem[{Buys and Blunsom(2017)}]{jan2017}
Jan Buys and Phil Blunsom. 2017.
\newblock \href {https://doi.org/10.18653/v1/P17-1112} {Robust incremental
  neural semantic graph parsing}.
\newblock In \emph{Proceedings of the 55th Annual Meeting of the Association
  for Computational Linguistics (Volume 1: Long Papers)}, pages 1215--1226,
  Vancouver, Canada. Association for Computational Linguistics.

\bibitem[{Chandurkar and Bansal(2017)}]{chandurkar2017ir}
Avani Chandurkar and Ajay Bansal. 2017.
\newblock \href {https://ieeexplore.ieee.org/abstract/document/7889571}
  {Information retrieval from a structured knowledgebase}.
\newblock In \emph{11th {IEEE} International Conference on Semantic Computing,
  {ICSC} 2017, San Diego, CA, USA, January 30 - February 1, 2017}, pages
  407--412. {IEEE} Computer Society.

\bibitem[{Chang and Lu(2021)}]{chang2021rethinking}
Ting-Yun Chang and Chi-Jen Lu. 2021.
\newblock \href {https://aclanthology.org/2021.findings-emnlp.61} {Rethinking
  why intermediate-task fine-tuning works}.
\newblock In \emph{Findings of the Association for Computational Linguistics:
  EMNLP 2021}, pages 706--713. Association for Computational Linguistics.

\bibitem[{Che et~al.(2019)Che, Dou, Xu, Wang, Liu, and Liu}]{che2019pipeline}
Wanxiang Che, Longxu Dou, Yang Xu, Yuxuan Wang, Yijia Liu, and Ting Liu. 2019.
\newblock \href {https://doi.org/10.18653/v1/K19-2007} {{HIT}-{SCIR} at {MRP}
  2019: A unified pipeline for meaning representation parsing via efficient
  training and effective encoding}.
\newblock In \emph{Proceedings of the Shared Task on Cross-Framework Meaning
  Representation Parsing at the 2019 Conference on Natural Language Learning},
  pages 76--85, Hong Kong. Association for Computational Linguistics.

\bibitem[{Che et~al.(2012)Che, Zhang, Shao, and Liu}]{che2016semeval}
Wanxiang Che, Meishan Zhang, Yanqiu Shao, and Ting Liu. 2012.
\newblock \href {https://aclanthology.org/S12-1050} {{S}em{E}val-2012 task 5:
  {C}hinese semantic dependency parsing}.
\newblock In \emph{*{SEM} 2012: The First Joint Conference on Lexical and
  Computational Semantics {--} Volume 1: Proceedings of the main conference and
  the shared task, and Volume 2: Proceedings of the Sixth International
  Workshop on Semantic Evaluation ({S}em{E}val 2012)}, pages 378--384,
  Montr{\'e}al, Canada. Association for Computational Linguistics.

\bibitem[{Clark et~al.(2019)Clark, Khandelwal, Levy, and
  Manning}]{clark2019bert}
Kevin Clark, Urvashi Khandelwal, Omer Levy, and Christopher~D. Manning. 2019.
\newblock \href {https://doi.org/10.18653/v1/W19-4828} {What does {BERT} look
  at? an analysis of {BERT}{'}s attention}.
\newblock In \emph{Proceedings of the 2019 ACL Workshop BlackboxNLP: Analyzing
  and Interpreting Neural Networks for NLP}, pages 276--286, Florence, Italy.
  Association for Computational Linguistics.

\bibitem[{Clark et~al.(2018)Clark, Luong, Manning, and Le}]{clark2018cvt}
Kevin Clark, Minh-Thang Luong, Christopher~D. Manning, and Quoc Le. 2018.
\newblock \href {https://doi.org/10.18653/v1/D18-1217} {Semi-supervised
  sequence modeling with cross-view training}.
\newblock In \emph{Proceedings of the 2018 Conference on Empirical Methods in
  Natural Language Processing}, pages 1914--1925, Brussels, Belgium.
  Association for Computational Linguistics.

\bibitem[{Devlin et~al.(2019)Devlin, Chang, Lee, and
  Toutanova}]{devlin2019bert}
Jacob Devlin, Ming-Wei Chang, Kenton Lee, and Kristina Toutanova. 2019.
\newblock \href {https://doi.org/10.18653/v1/N19-1423} {{BERT}: Pre-training of
  deep bidirectional transformers for language understanding}.
\newblock In \emph{Proceedings of the 2019 Conference of the North {A}merican
  Chapter of the Association for Computational Linguistics: Human Language
  Technologies, Volume 1 (Long and Short Papers)}, pages 4171--4186,
  Minneapolis, Minnesota. Association for Computational Linguistics.

\bibitem[{Dozat and Manning(2017)}]{dozat2017biaffine}
Timothy Dozat and Christopher~D. Manning. 2017.
\newblock \href {https://openreview.net/forum?id=Hk95PK9le} {Deep biaffine
  attention for neural dependency parsing}.
\newblock In \emph{5th International Conference on Learning Representations,
  {ICLR} 2017, Toulon, France, April 24-26, 2017, Conference Track
  Proceedings}. OpenReview.net.

\bibitem[{Dozat and Manning(2018)}]{dozat2018sedp}
Timothy Dozat and Christopher~D. Manning. 2018.
\newblock \href {https://doi.org/10.18653/v1/P18-2077} {Simpler but more
  accurate semantic dependency parsing}.
\newblock In \emph{Proceedings of the 56th Annual Meeting of the Association
  for Computational Linguistics (Volume 2: Short Papers)}, pages 484--490,
  Melbourne, Australia. Association for Computational Linguistics.

\bibitem[{Fern{\'a}ndez-Gonz{\'a}lez and
  G{\'o}mez-Rodr{\'\i}guez(2020)}]{fernandez2020transition}
Daniel Fern{\'a}ndez-Gonz{\'a}lez and Carlos G{\'o}mez-Rodr{\'\i}guez. 2020.
\newblock \href {https://doi.org/10.18653/v1/2020.acl-main.629}
  {Transition-based semantic dependency parsing with pointer networks}.
\newblock In \emph{Proceedings of the 58th Annual Meeting of the Association
  for Computational Linguistics}, pages 7035--7046, Online. Association for
  Computational Linguistics.

\bibitem[{Gan et~al.(2021)Gan, Meng, Kuang, Sun, Fan, Wu, and Li}]{gan21mrc}
Leilei Gan, Yuxing Meng, Kun Kuang, Xiaofei Sun, Chun Fan, Fei Wu, and Jiwei
  Li. 2021.
\newblock \href {https://arxiv.org/abs/2105.07654} {Dependency parsing as
  mrc-based span-span prediction}.
\newblock \emph{ArXiv preprint}, abs/2105.07654.

\bibitem[{Gardner et~al.(2018)Gardner, Grus, Neumann, Tafjord, Dasigi, Liu,
  Peters, Schmitz, and Zettlemoyer}]{gardner2018allennlp}
Matt Gardner, Joel Grus, Mark Neumann, Oyvind Tafjord, Pradeep Dasigi,
  Nelson~F. Liu, Matthew Peters, Michael Schmitz, and Luke Zettlemoyer. 2018.
\newblock \href {https://doi.org/10.18653/v1/W18-2501} {{A}llen{NLP}: A deep
  semantic natural language processing platform}.
\newblock In \emph{Proceedings of Workshop for {NLP} Open Source Software
  ({NLP}-{OSS})}, pages 1--6, Melbourne, Australia. Association for
  Computational Linguistics.

\bibitem[{Geva et~al.(2020)Geva, Gupta, and Berant}]{geva2020injecting}
Mor Geva, Ankit Gupta, and Jonathan Berant. 2020.
\newblock \href {https://doi.org/10.18653/v1/2020.acl-main.89} {Injecting
  numerical reasoning skills into language models}.
\newblock In \emph{Proceedings of the 58th Annual Meeting of the Association
  for Computational Linguistics}, pages 946--958, Online. Association for
  Computational Linguistics.

\bibitem[{He and D.~Choi(2020)}]{he2020bert}
Han He and Jinho D.~Choi. 2020.
\newblock \href
  {https://aaai.org/ocs/index.php/FLAIRS/FLAIRS20/paper/view/18438}
  {Establishing strong baselines for the new decade: Sequence tagging,
  syntactic and semantic parsing with {BERT}}.
\newblock In \emph{Proceedings of the Thirty-Third International Florida
  Artificial Intelligence Research Society Conference, Originally to be held in
  North Miami Beach, Florida, USA, May 17-20, 2020}, pages 228--233. {AAAI}
  Press.

\bibitem[{Hewitt and Manning(2019)}]{hewitt2019structural}
John Hewitt and Christopher~D. Manning. 2019.
\newblock \href {https://doi.org/10.18653/v1/N19-1419} {{A} structural probe
  for finding syntax in word representations}.
\newblock In \emph{Proceedings of the 2019 Conference of the North {A}merican
  Chapter of the Association for Computational Linguistics: Human Language
  Technologies, Volume 1 (Long and Short Papers)}, pages 4129--4138,
  Minneapolis, Minnesota. Association for Computational Linguistics.

\bibitem[{Kudo and Richardson(2018)}]{kudo2018sentencepiece}
Taku Kudo and John Richardson. 2018.
\newblock \href {https://doi.org/10.18653/v1/D18-2012} {{S}entence{P}iece: A
  simple and language independent subword tokenizer and detokenizer for neural
  text processing}.
\newblock In \emph{Proceedings of the 2018 Conference on Empirical Methods in
  Natural Language Processing: System Demonstrations}, pages 66--71, Brussels,
  Belgium. Association for Computational Linguistics.

\bibitem[{Lewis et~al.(2020)Lewis, Liu, Goyal, Ghazvininejad, Mohamed, Levy,
  Stoyanov, and Zettlemoyer}]{lewis2020bart}
Mike Lewis, Yinhan Liu, Naman Goyal, Marjan Ghazvininejad, Abdelrahman Mohamed,
  Omer Levy, Veselin Stoyanov, and Luke Zettlemoyer. 2020.
\newblock \href {https://doi.org/10.18653/v1/2020.acl-main.703} {{BART}:
  Denoising sequence-to-sequence pre-training for natural language generation,
  translation, and comprehension}.
\newblock In \emph{Proceedings of the 58th Annual Meeting of the Association
  for Computational Linguistics}, pages 7871--7880, Online. Association for
  Computational Linguistics.

\bibitem[{Li et~al.(2019)Li, Peng, Zhang, Wang, and Si}]{li2019codt}
Zhenghua Li, Xue Peng, Min Zhang, Rui Wang, and Luo Si. 2019.
\newblock \href {https://doi.org/10.18653/v1/P19-1229} {Semi-supervised domain
  adaptation for dependency parsing}.
\newblock In \emph{Proceedings of the 57th Annual Meeting of the Association
  for Computational Linguistics}, pages 2386--2395, Florence, Italy.
  Association for Computational Linguistics.

\bibitem[{Li et~al.(2018)Li, Cai, He, and Zhao}]{li2018seq2seq}
Zuchao Li, Jiaxun Cai, Shexia He, and Hai Zhao. 2018.
\newblock \href {https://aclanthology.org/C18-1271} {Seq2seq dependency
  parsing}.
\newblock In \emph{Proceedings of the 27th International Conference on
  Computational Linguistics}, pages 3203--3214, Santa Fe, New Mexico, USA.
  Association for Computational Linguistics.

\bibitem[{Lin et~al.(2021)Lin, Li, Li, and Luo}]{lin2021unsupervised}
Boda Lin, Mingzheng Li, Si~Li, and Yong Luo. 2021.
\newblock \href {https://aclanthology.org/2021.findings-emnlp.186}
  {Unsupervised domain adaptation method with semantic-structural alignment for
  dependency parsing}.
\newblock In \emph{Findings of the Association for Computational Linguistics:
  EMNLP 2021}, pages 2158--2167, Punta Cana, Dominican Republic. Association
  for Computational Linguistics.

\bibitem[{Liu et~al.(2019)Liu, Ott, Goyal, Du, Joshi, Chen, Levy, Lewis,
  Zettlemoyer, and Stoyanov}]{liu2019roberta}
Yinhan Liu, Myle Ott, Naman Goyal, Jingfei Du, Mandar Joshi, Danqi Chen, Omer
  Levy, Mike Lewis, Luke Zettlemoyer, and Veselin Stoyanov. 2019.
\newblock \href {https://arxiv.org/abs/1907.11692} {Roberta: A robustly
  optimized bert pretraining approach}.
\newblock \emph{ArXiv preprint}, abs/1907.11692.

\bibitem[{Loshchilov and Hutter(2019)}]{loshchilov2017adamw}
Ilya Loshchilov and Frank Hutter. 2019.
\newblock \href {https://openreview.net/forum?id=Bkg6RiCqY7} {Decoupled weight
  decay regularization}.
\newblock In \emph{7th International Conference on Learning Representations,
  {ICLR} 2019, New Orleans, LA, USA, May 6-9, 2019}. OpenReview.net.

\bibitem[{Ma et~al.(2018)Ma, Hu, Liu, Peng, Neubig, and Hovy}]{ma2018stackptr}
Xuezhe Ma, Zecong Hu, Jingzhou Liu, Nanyun Peng, Graham Neubig, and Eduard
  Hovy. 2018.
\newblock \href {https://doi.org/10.18653/v1/P18-1130} {Stack-pointer networks
  for dependency parsing}.
\newblock In \emph{Proceedings of the 56th Annual Meeting of the Association
  for Computational Linguistics (Volume 1: Long Papers)}, pages 1403--1414,
  Melbourne, Australia. Association for Computational Linguistics.

\bibitem[{Manning et~al.(2014)Manning, Surdeanu, Bauer, Finkel, Bethard, and
  McClosky}]{manning2014stanford}
Christopher Manning, Mihai Surdeanu, John Bauer, Jenny Finkel, Steven Bethard,
  and David McClosky. 2014.
\newblock \href {https://doi.org/10.3115/v1/P14-5010} {The {S}tanford
  {C}ore{NLP} natural language processing toolkit}.
\newblock In \emph{Proceedings of 52nd Annual Meeting of the Association for
  Computational Linguistics: System Demonstrations}, pages 55--60, Baltimore,
  Maryland. Association for Computational Linguistics.

\bibitem[{Marcus et~al.(1993)Marcus, Santorini, and
  Marcinkiewicz}]{marcus1993ptb3}
Mitchell~P. Marcus, Beatrice Santorini, and Mary~Ann Marcinkiewicz. 1993.
\newblock \href {https://aclanthology.org/J93-2004} {Building a large annotated
  corpus of {E}nglish: The {P}enn {T}reebank}.
\newblock \emph{Computational Linguistics}, 19(2):313--330.

\bibitem[{Oepen et~al.(2014)Oepen, Kuhlmann, Miyao, Zeman, Flickinger,
  Haji{\v{c}}, Ivanova, and Zhang}]{stephan2015sdp15}
Stephan Oepen, Marco Kuhlmann, Yusuke Miyao, Daniel Zeman, Dan Flickinger, Jan
  Haji{\v{c}}, Angelina Ivanova, and Yi~Zhang. 2014.
\newblock \href {https://doi.org/10.3115/v1/S14-2008} {{S}em{E}val 2014 task 8:
  Broad-coverage semantic dependency parsing}.
\newblock In \emph{Proceedings of the 8th International Workshop on Semantic
  Evaluation ({S}em{E}val 2014)}, pages 63--72, Dublin, Ireland. Association
  for Computational Linguistics.

\bibitem[{Peng et~al.(2017)Peng, Thomson, and Smith}]{peng2017multi}
Hao Peng, Sam Thomson, and Noah~A. Smith. 2017.
\newblock \href {https://doi.org/10.18653/v1/P17-1186} {Deep multitask learning
  for semantic dependency parsing}.
\newblock In \emph{Proceedings of the 55th Annual Meeting of the Association
  for Computational Linguistics (Volume 1: Long Papers)}, pages 2037--2048,
  Vancouver, Canada. Association for Computational Linguistics.

\bibitem[{Peng et~al.(2019)Peng, Li, Zhang, Rui, Zhang, and Si}]{peng2019nlpcc}
Xue Peng, Zhenghua Li, Min Zhang, Wang Rui, Yue Zhang, and Luo Si. 2019.
\newblock \href
  {https://link.springer.com/chapter/10.1007/978-3-030-32236-6_69} {Overview of
  the nlpcc 2019 shared task: Cross-domain dependency parsing}.
\newblock In \emph{Natural Language Processing and Chinese Computing - 8th
  {CCF} International Conference, {NLPCC} 2019, Dunhuang, China, October 9-14,
  2019, Proceedings, Part {II}}, volume 11839, pages 760--771. Springer.

\bibitem[{Pruksachatkun et~al.(2020)Pruksachatkun, Phang, Liu, Htut, Zhang,
  Pang, Vania, Kann, and Bowman}]{pruksachatkun2020intermediate}
Yada Pruksachatkun, Jason Phang, Haokun Liu, Phu~Mon Htut, Xiaoyi Zhang,
  Richard~Yuanzhe Pang, Clara Vania, Katharina Kann, and Samuel~R. Bowman.
  2020.
\newblock \href {https://doi.org/10.18653/v1/2020.acl-main.467}
  {Intermediate-task transfer learning with pretrained language models: When
  and why does it work?}
\newblock In \emph{Proceedings of the 58th Annual Meeting of the Association
  for Computational Linguistics}, pages 5231--5247, Online. Association for
  Computational Linguistics.

\bibitem[{Raffel et~al.(2020)Raffel, Shazeer, Roberts, Lee, Narang, Matena,
  Zhou, Li, and Liu}]{colin2020t5}
Colin Raffel, Noam Shazeer, Adam Roberts, Katherine Lee, Sharan Narang, Michael
  Matena, Yanqi Zhou, Wei Li, and Peter~J. Liu. 2020.
\newblock \href {http://jmlr.org/papers/v21/20-074.html} {Exploring the limits
  of transfer learning with a unified text-to-text transformer}.
\newblock \emph{J. Mach. Learn. Res.}, 21:140:1--140:67.

\bibitem[{Strzyz et~al.(2019)Strzyz, Vilares, and
  G{\'o}mez-Rodr{\'\i}guez}]{strzyz2019viable}
Michalina Strzyz, David Vilares, and Carlos G{\'o}mez-Rodr{\'\i}guez. 2019.
\newblock \href {https://doi.org/10.18653/v1/N19-1077} {Viable dependency
  parsing as sequence labeling}.
\newblock In \emph{Proceedings of the 2019 Conference of the North {A}merican
  Chapter of the Association for Computational Linguistics: Human Language
  Technologies, Volume 1 (Long and Short Papers)}, pages 717--723, Minneapolis,
  Minnesota. Association for Computational Linguistics.

\bibitem[{Teney et~al.(2017)Teney, Liu, and van~den Hengel}]{teney2017qa}
Damien Teney, Lingqiao Liu, and Anton van~den Hengel. 2017.
\newblock \href {https://doi.org/10.1109/CVPR.2017.344} {Graph-structured
  representations for visual question answering}.
\newblock In \emph{2017 {IEEE} Conference on Computer Vision and Pattern
  Recognition, {CVPR} 2017, Honolulu, HI, USA, July 21-26, 2017}, pages
  3233--3241. {IEEE} Computer Society.

\bibitem[{Vacareanu et~al.(2020)Vacareanu, Gouveia~Barbosa,
  Valenzuela-Esc{\'a}rcega, and Surdeanu}]{vacareanu2020pat}
Robert Vacareanu, George~Caique Gouveia~Barbosa, Marco~A.
  Valenzuela-Esc{\'a}rcega, and Mihai Surdeanu. 2020.
\newblock \href {https://aclanthology.org/2020.lrec-1.643} {Parsing as
  tagging}.
\newblock In \emph{Proceedings of the 12th Language Resources and Evaluation
  Conference}, pages 5225--5231, Marseille, France. European Language Resources
  Association.

\bibitem[{Vaswani et~al.(2017)Vaswani, Shazeer, Parmar, Uszkoreit, Jones,
  Gomez, Kaiser, and Polosukhin}]{vaswani2017attention}
Ashish Vaswani, Noam Shazeer, Niki Parmar, Jakob Uszkoreit, Llion Jones,
  Aidan~N. Gomez, Lukasz Kaiser, and Illia Polosukhin. 2017.
\newblock \href
  {https://proceedings.neurips.cc/paper/2017/hash/3f5ee243547dee91fbd053c1c4a845aa-Abstract.html}
  {Attention is all you need}.
\newblock In \emph{Advances in Neural Information Processing Systems 30: Annual
  Conference on Neural Information Processing Systems 2017, December 4-9, 2017,
  Long Beach, CA, {USA}}, pages 5998--6008.

\bibitem[{Wang and Tu(2020)}]{wang2020second}
Xinyu Wang and Kewei Tu. 2020.
\newblock \href {https://aclanthology.org/2020.aacl-main.12} {Second-order
  neural dependency parsing with message passing and end-to-end training}.
\newblock In \emph{Proceedings of the 1st Conference of the Asia-Pacific
  Chapter of the Association for Computational Linguistics and the 10th
  International Joint Conference on Natural Language Processing}, pages 93--99,
  Suzhou, China. Association for Computational Linguistics.

\bibitem[{Wang et~al.(2018)Wang, Che, Guo, and Liu}]{wang2019sedp}
Yuxuan Wang, Wanxiang Che, Jiang Guo, and Ting Liu. 2018.
\newblock \href
  {https://www.aaai.org/ocs/index.php/AAAI/AAAI18/paper/view/16549} {A neural
  transition-based approach for semantic dependency graph parsing}.
\newblock In \emph{Proceedings of the Thirty-Second {AAAI} Conference on
  Artificial Intelligence, (AAAI-18), the 30th innovative Applications of
  Artificial Intelligence (IAAI-18), and the 8th {AAAI} Symposium on
  Educational Advances in Artificial Intelligence (EAAI-18), New Orleans,
  Louisiana, USA, February 2-7, 2018}, pages 5561--5568. {AAAI} Press.

\bibitem[{Wolf et~al.(2020)Wolf, Debut, Sanh, Chaumond, Delangue, Moi, Cistac,
  Rault, Louf, Funtowicz, Davison, Shleifer, von Platen, Ma, Jernite, Plu, Xu,
  Le~Scao, Gugger, Drame, Lhoest, and Rush}]{wolf2020transformers}
Thomas Wolf, Lysandre Debut, Victor Sanh, Julien Chaumond, Clement Delangue,
  Anthony Moi, Pierric Cistac, Tim Rault, Remi Louf, Morgan Funtowicz, Joe
  Davison, Sam Shleifer, Patrick von Platen, Clara Ma, Yacine Jernite, Julien
  Plu, Canwen Xu, Teven Le~Scao, Sylvain Gugger, Mariama Drame, Quentin Lhoest,
  and Alexander Rush. 2020.
\newblock \href {https://doi.org/10.18653/v1/2020.emnlp-demos.6} {Transformers:
  State-of-the-art natural language processing}.
\newblock In \emph{Proceedings of the 2020 Conference on Empirical Methods in
  Natural Language Processing: System Demonstrations}, pages 38--45, Online.
  Association for Computational Linguistics.

\bibitem[{Xue et~al.(2021)Xue, Constant, Roberts, Kale, Al-Rfou, Siddhant,
  Barua, and Raffel}]{xue2021mt5}
Linting Xue, Noah Constant, Adam Roberts, Mihir Kale, Rami Al-Rfou, Aditya
  Siddhant, Aditya Barua, and Colin Raffel. 2021.
\newblock \href {https://doi.org/10.18653/v1/2021.naacl-main.41} {m{T}5: A
  massively multilingual pre-trained text-to-text transformer}.
\newblock In \emph{Proceedings of the 2021 Conference of the North American
  Chapter of the Association for Computational Linguistics: Human Language
  Technologies}, pages 483--498, Online. Association for Computational
  Linguistics.

\bibitem[{Yu et~al.(2015)Yu, Elkaref, and Bohnet}]{yu2015self}
Juntao Yu, Mohab Elkaref, and Bernd Bohnet. 2015.
\newblock \href {https://aclanthology.org/W15-2201/} {Domain adaptation for
  dependency parsing via self-trainging}.
\newblock In \emph{Proceedings of the 14th International Conference on Parsing
  Technologies}, pages 1--10, Bilbao, Spain. Association for Computational
  Linguistics.

\bibitem[{Zhang et~al.(2019)Zhang, Ma, Duh, and Van~Durme}]{zhang2019broad}
Sheng Zhang, Xutai Ma, Kevin Duh, and Benjamin Van~Durme. 2019.
\newblock \href {https://doi.org/10.18653/v1/D19-1392} {Broad-coverage semantic
  parsing as transduction}.
\newblock In \emph{Proceedings of the 2019 Conference on Empirical Methods in
  Natural Language Processing and the 9th International Joint Conference on
  Natural Language Processing (EMNLP-IJCNLP)}, pages 3786--3798, Hong Kong,
  China. Association for Computational Linguistics.

\bibitem[{Zhang et~al.(2016)Zhang, Du, Sun, and Wan}]{zhang2016}
Xun Zhang, Yantao Du, Weiwei Sun, and Xiaojun Wan. 2016.
\newblock \href {https://doi.org/10.1162/COLI_a_00252} {Transition-based
  parsing for deep dependency structures}.
\newblock \emph{Computational Linguistics}, 42(3):353--389.

\bibitem[{Zhang et~al.(2020)Zhang, Li, and Zhang}]{li2020crf}
Yu~Zhang, Zhenghua Li, and Min Zhang. 2020.
\newblock \href {https://doi.org/10.18653/v1/2020.acl-main.302} {Efficient
  second-order {T}ree{CRF} for neural dependency parsing}.
\newblock In \emph{Proceedings of the 58th Annual Meeting of the Association
  for Computational Linguistics}, pages 3295--3305, Online. Association for
  Computational Linguistics.

\end{thebibliography}
\clearpage

\appendix

\section{Dataset Statistics}
\label{app:data}
\begin{table}[t]
    \centering
    \scalebox{0.8}{
        \begin{tabular}{ccccc}
        \toprule
        Set & Section & Sentences & Words \\
        \midrule
        Train & [$2$-$21$] & $39,832$ & $95,0028$ \\
        Dev & [$22$] & $1,700$ & $40,117$ \\
        Test & [$23$] & $2,416$ & $56,684$ \\
        \bottomrule
        \end{tabular}
    }
    \caption{Data statistics of PTB.}
    \label{tab:exp:data:ptb}
\end{table}
\begin{table}[t]
    \centering
    \scalebox{0.8}{
        \begin{tabular}{ccccc}
        \toprule
        Domain & Train Set & Dev Set & Test Set & Unlabeled Set \\
        \midrule
        BC & $16.3$K & $1$K & $2$K & \makecell[c]{--} \\
        PB & $5.1$K & $1.3$K & $2.6$K & $291$K\\
        PC & $6.6$K & $1.3$K & $2.6$K & $349$K\\
        ZX & $1.6$K & $0.5$K & $1.1$K & $33$K\\
        \bottomrule
        \end{tabular}
    }
    \caption{Data statistics of CODT.}
    \label{tab:exp:data:codt}
\end{table}
\begin{table}[t]
    \centering
    \scalebox{0.8}{
        \begin{tabular}{ccccc}
        \toprule
        Schema & Train Set & ID Test Set & OOD Test Set \\
        \midrule
        DM & $35,656$ & $1,410$ & $1,849$ \\
        PAS & $35,656$ & $1,410$ & $1,849$ \\
        PSD & $35,656$ & $1,410$ & $1,849$ \\
        \bottomrule
        \end{tabular}
    }
    \caption{Data statistics of SDP15.}
    \label{tab:exp:data:sdp15}
\end{table}
\begin{table}[t]
    \centering
    \scalebox{0.8}{
        \begin{tabular}{ccccc}
        \toprule
        Domain & Train Set & Dev Set & Test Set \\
        \midrule
        NEWS & $8,301$ & $534$ & $1,233$ \\
        TEXT & $128,095$ & $1,546$ & $3,096$ \\
        \bottomrule
        \end{tabular}
    }
    \caption{Data statistics of SemEval16.}
    \label{tab:exp:data:semeval16}
\end{table}
The details about the statistics of datasets used in this paper are shown on Table~\ref{tab:exp:data:ptb}, Table~\ref{tab:exp:data:codt}, Table~\ref{tab:exp:data:sdp15} and Table~\ref{tab:exp:data:semeval16}.

\section{More Details on Baseline}
\label{app:baseline}

% 补充具体的trick介绍
\vpara{Baselines for in-domain SyDP.}
\begin{itemize}[leftmargin=*]
\setlength\parskip{0pt}
\setlength\itemsep{3pt}
    \item [*]\footnote{* means model without PLM} \vpara{Biaffine:}~\citet{dozat2017biaffine} adopt biaffine attention mechanism into the graph-based method of dependency parsing.
    \item [*] \vpara{StackPTR:}~\citet{ma2018stackptr} introduce the pointer network into the transition-based methods of dependency parsing.
    \item [*] \vpara{CRF:}~\citet{li2020crf} improve the CRF to capture more high-order information in dependency parsing.
    % and achieve the SOTA performance among the parser without PLM.
    \item \footnote{$\bullet$ means sequence-based methods}\vpara{SeqNMT:}~\citet{li2018seq2seq} use an Encoder-Decoder architecture to achieve the Seq2Seq dependency parsing by sequence tagging. The BPE segmentation from Neural Machine Translation (NMT) and character embedding from AllenNLP~\cite{gardner2018allennlp} are applied to argument their model.
    \item \vpara{SeqViable:}~\citet{strzyz2019viable} explore four encodings of dependency trees and improve the performance comparing with~\citet{li2018seq2seq}.
    \item \vpara{PaT:}~\citet{vacareanu2020pat} use a simple tagging structure over BERT-base to achieve sequence labeling of dependency parsing.
    \item [+]\footnote{+ means model utilizing PLM} \vpara{CVT:}~\citet{clark2018cvt} propose another pre-train method named cross-view training, which can be used in many sequence constructing task including SyDP. The best results of CVT is achieved by the multi-task pre-training of SyDP and part-of-speech tagging.
    \item [+] \vpara{MP2O:}~\citet{wang2020second} use message passing GNN based on BERT to capture second-order information in SyDP.
    \item [+] \vpara{MRC:}~\citet{gan21mrc} use span-based method to construct the edges at the subtree level. The Machine Reading Comprehension (MRC) is applied to link the different span. RoBERTa-large~\cite{liu2019roberta} is applied to enhance the representation of parser.
\end{itemize}

\begin{figure*}[t]
\centering
    \includegraphics[width=1.0\linewidth]{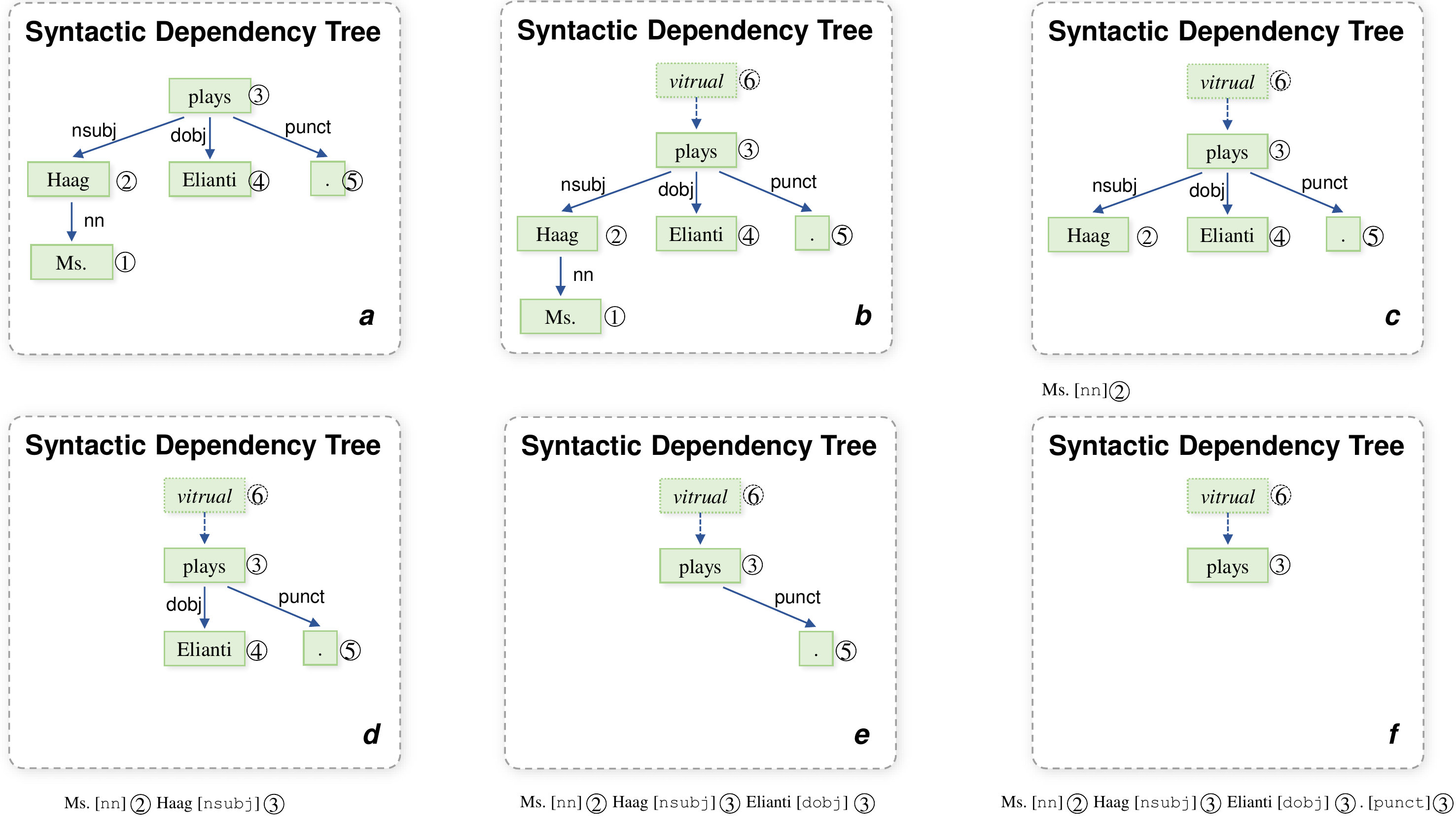}
\caption{
The Prufer Sequence of sentence ``\textit{Ms. Haag plays Elianti .}'' is constructed from \textbf{\textit{a}} to \textbf{\textit{f}}.
}
\label{fig:prufer:construction}
\end{figure*}

\begin{figure}[t]
\centering
    \includegraphics[width=1.0\linewidth]{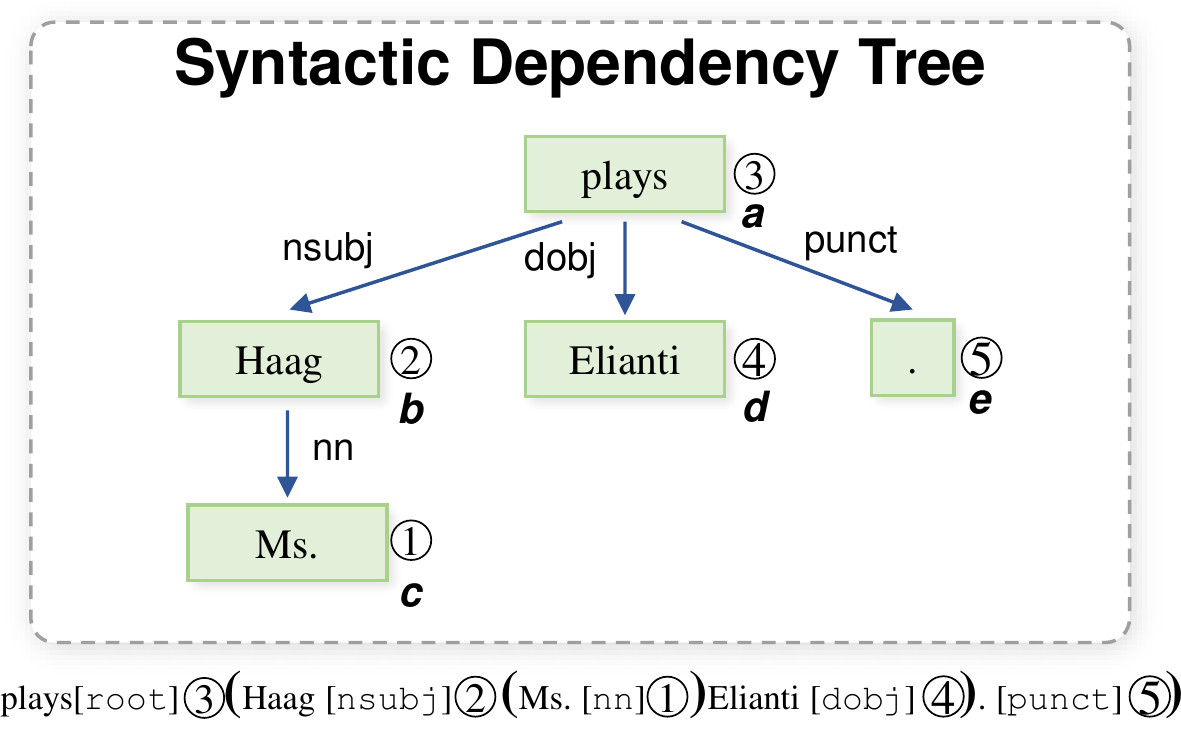}
\caption{
The Bracket Tree Sequence of sentence ``\textit{Ms. Haag plays Elianti .}'' is constructed following the topological order from \textit{\textbf{a}} to \textit{\textbf{e}}.
}
\label{fig:bracket:construction}
\end{figure}

\vpara{Baselines for cross-domain SyDP.}
\begin{itemize}[leftmargin=*]
\setlength\parskip{0pt}
\setlength\itemsep{3pt}
    \item [*] \vpara{Biaffine:}~\citet{peng2019nlpcc,li2019codt} use Biaffine trained on source domain and test on target domain as the baseline of unsupervised cross-domain SyDP.
    \item [*] \vpara{SSADP:}~\citet{lin2021unsupervised} use both semantic and structural feature to achieve the domain adaptation of unsupervised cross-domain parsing.
    \item [+] \vpara{ELMo:}~\citet{li2019codt} use ELMo with intermediate fine-tuning in unlabeled text of target domain to achieve the SOTA on unsupervised cross-domain SyDP.
\end{itemize}

\vpara{Baselines for SeDP.}
\begin{itemize}[leftmargin=*]
\setlength\parskip{0pt}
\setlength\itemsep{3pt}
    \item [*] \vpara{Biaffine:}~\citet{dozat2018sedp} transfer the Biaffine model from SyDP to SeDP.
    \item [*] \vpara{BS-IT:}~\citet{wang2019sedp} use graph-based method for SeDP.
    \item \vpara{HIT-SCIR:}~\citet{che2019pipeline} propose a BERT-based pipeline model for SeDP.
    \item \vpara{BERT+Flair:}~\citet{he2020bert} use BERT and flair embedding~\cite{akbik2018flair} to argument their modificated Biaffine.
\end{itemize}

\section{Construction of Prufer Sequence}
\label{app:prufer}

\subsection{Prufer Sequence}
% We first describe the construction of original prufer sequence from a simple sample.

% There is a labeled undirect tree as shown in Fig~\ref{somefig}.
The principle of construction is deleting the leaf node with minimum index and adding the index of its farther node into the prufer sequence.
This process is repeated more times until there are only two nodes left in the tree.

\subsection{Prufer for Parsing Tree}
The arc in parsing tree is directed and thus is a rooted tree.
% This may leads to the root node with smaller index may by deleted prematurely.
When all the son nodes with smaller index are deleted, the root node will be treated as a leaf node then deleted in the next step.
To address this problem, we add a virtual node with the maximum index and build a arc from virtual node to the real root.
This virtual root force the root node always being a leaf node in the whole construction of prufer sequence.
The overall construction process as shown on Figure~\ref{fig:prufer:construction} (a)\textasciitilde(f).

% The final Prufer Sequence is like:

% \begin{equation*}
% \Big(...\Big[\mathtt{SPT}\Big]\ \underbrace{w_i\ \Big[\mathtt{REL}\left(r_i^j\right)\Big]\ \mathtt{POS}\left(h_i^j\right)}_{\text{\small one dependency unit}}\ \Big[\mathtt{SPT}\Big]...\Big)
% \end{equation*}

\begin{figure}[t]
\centering
    \includegraphics[width=1.0\linewidth]{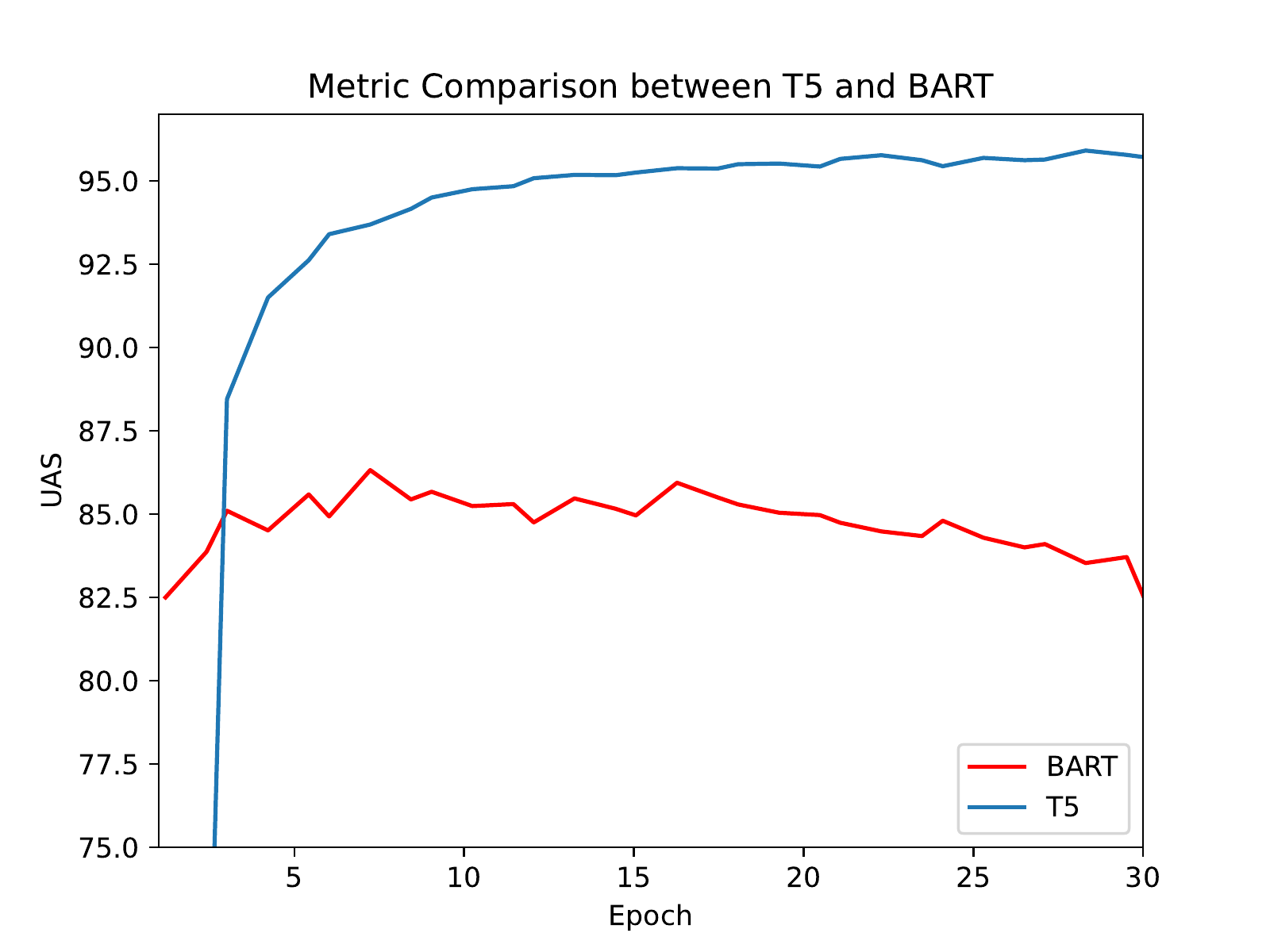}
\caption{
The UAS curves on dev sets of PTB between of T5 and BART.
}
\label{fig:bart}
\end{figure}

\section{Construction of Bracket Tree}
\label{app:bracket}
The Bracket Tree uses \textit{Bracket} to indicate levels of nodes.
All the nodes belonging to the same level are wrapped in the same pair of brackets.
The process of construction is shown on Figure~\ref{fig:bracket:construction}.

\section{Comparison between T5 and BART}
\label{app:bart}
Figure~\ref{fig:bart} shows the UAS comparison on dev sets of PTB between the T5 and BART in first 30 epochs.
After the first two epochs, the performance of T5 raise rapidly and can better maintain performance in the later stages of training.
Although BART achieves a better performance in the first two round, but there is not much room for performance improvement.
To make matters worse, it can be clearly seen that after achieving the best performance, BART is very unstable, and even a significant performance drop has occurred.

\end{document}